\documentclass{article}

\usepackage{arxiv}

\usepackage[utf8]{inputenc} 
\usepackage[T1]{fontenc}    
\usepackage{hyperref}       
\usepackage{url}            
\usepackage{booktabs}       
\usepackage{amsfonts}       
\usepackage{nicefrac}       
\usepackage{microtype}      
\usepackage{lipsum}
\usepackage{graphicx}

\usepackage{caption} 
\usepackage{algorithm}
\usepackage{algpseudocode}
\usepackage{amsmath}
\usepackage{amssymb}
\usepackage{enumitem}
\usepackage{array}
\usepackage{multirow}
\usepackage{tikz}

\usepackage{newfloat}
\usepackage{listings}
\newfloat{listing}{tb}{lst}{}
\floatname{listing}{Listing}
\newcommand{\new}[1]{#1}
\graphicspath{ {./images/} }

\title{Lingjing: A Simulation Testbed for Multi-Agent Embodied Tasks in Open-Ended Cities}

\author{
 Xiaohe Li, Yiru Wang, Junhao Fan, Mingyuan Liu, Jie Huang, Kaixin Zhang, Jiahao Li\\
 	the Key Laboratory of Target\\
 	Cognition and Application Technology\\
	Aerospace Information Research Institute\\
	Chinese Academy of Sciences \\
  \texttt{lixiaohe@aircas.ac.cn} \\
   \And
 Chen Qian \\
  School of Artificial Intelligence\\
  Shanghai Jiaotong University \\
  \texttt{qianc@sjtu.edu.cn} \\
  \And
 Zide Fan \\
the Key Laboratory of Target\\
Cognition and Application Technology\\
Aerospace Information Research Institute\\
Chinese Academy of Sciences\\
\texttt{fanzd@aircas.ac.cn}}

\begin{document}
\maketitle
\begin{abstract}
Urban embodied intelligence requires coordination among heterogeneous agents (e.g., UAVs, ground robots, and autonomous vehicles) in dynamic cities. Simulators therefore provide a scalable foundation for developing and evaluating such coordination. Existing platforms nevertheless isolate different embodiments and decouple them from task design and evaluation. We present \textbf{Lingjing}, a simulation platform for heterogeneous multi-agent embodied intelligence in open-ended urban environments. Lingjing reconstructs and renders evolving cities from geographic data, synchronizes multiple physics engines, and exposes shared physical and structured urban state to agents. Its Gym-like interface supports user-defined ReAct agents and single- or multi-agent natural-language missions with configurable star or broadcast communication and resource constraints. Each episode becomes an attribution-ready replay that links agent trajectories and communication to relation-graph changes, resource consumption, and engine-based evaluations for systematic diagnosis. We evaluate twelve vision-language models on nine urban tasks under a shared engine-in-the-loop protocol. Controlled studies further examine communication, scalability, robustness, and failure provenance. Results expose persistent bottlenecks in grounding and long-horizon execution. They also show task-dependent coordination trade-offs and diminishing returns from added capacity, while heavier workloads further reduce success. Lingjing provides a unified testbed that enables reproducible end-to-end evaluation and systematic failure diagnosis in urban multi-agent embodied intelligence.
\end{abstract}

\section{Introduction}
\label{sec:intro}
Multimodal large language models (MLLMs)~\cite{gpt4,team2023gemini,liu2024deepseek,bai2025qwen3} have accelerated the development of embodied agents, supporting tasks such as vision-language navigation (VLN), complex manipulation, and environmental inspection. Increasingly, the object of study is shifting from a single robot~\cite{liembodied,yangembodiedbench} operating in a static room toward coordinated operation across an entire city~\cite{gao2024embodiedcity, metaurban}. City-scale applications such as delivery, infrastructure inspection, and emergency response require heterogeneous mobile and fixed assets to coordinate across urban blocks. The challenge extends beyond perception and low-level control. For example, agents may need to adapt to traffic and weather disruptions under limited communication and resource budgets while accounting for how their decisions affect one another and the city.

\begin{figure}[t]
    \centering
    \includegraphics[width=\linewidth]{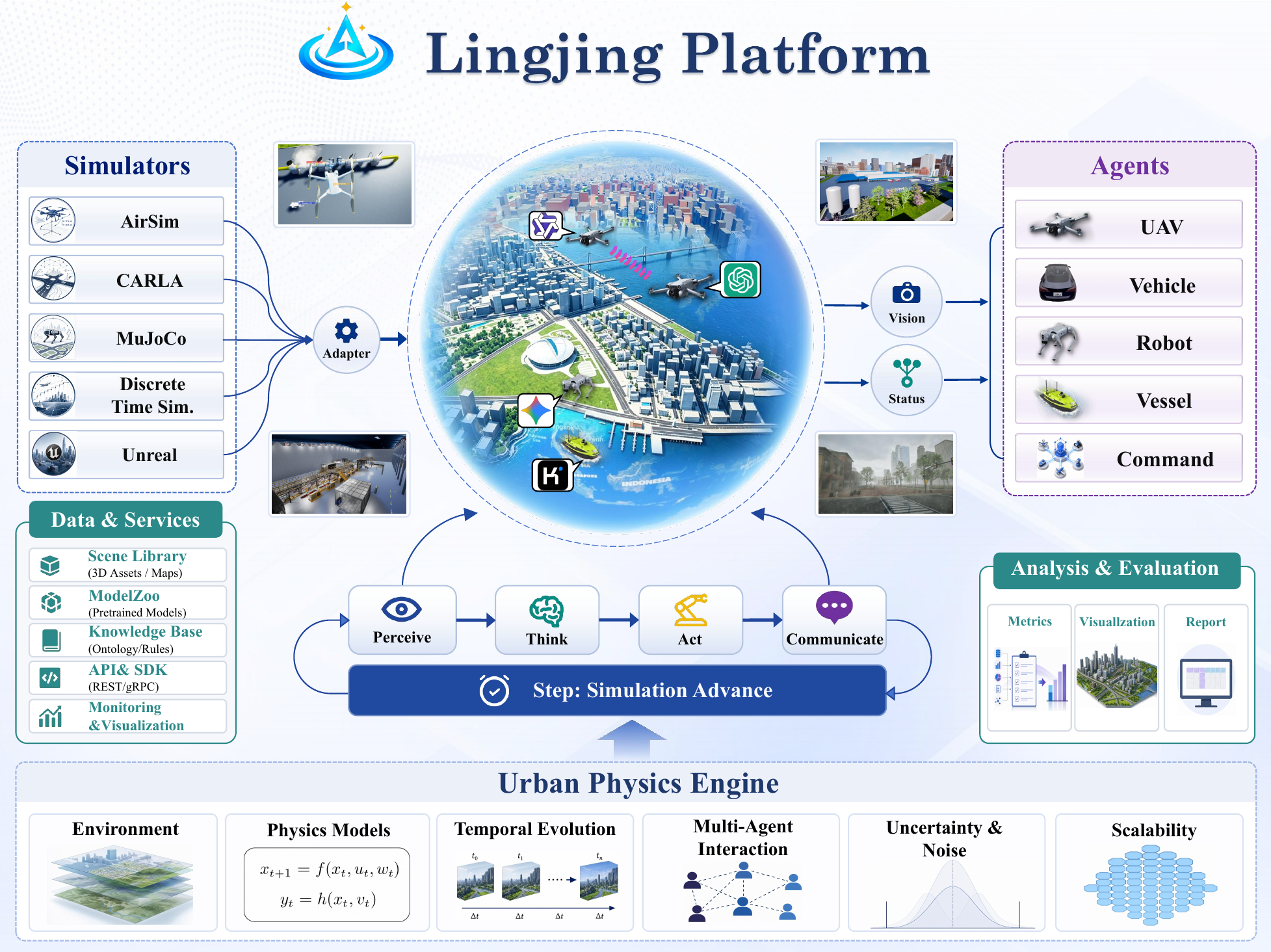}
    \caption{Overview of Lingjing. Its Urban Physics Engine integrates multiple simulators into a shared, evolving city environment for evaluating MLLM-driven heterogeneous agents across perception, reasoning, action, and communication.}
    \label{fig:lingjing-overview}
\end{figure}

Training and evaluating embodied intelligence directly in real cities is costly, slow, and safety-critical. Synchronized data from heterogeneous agents are difficult to collect, and urban incidents cannot be replayed to identify which agent, message, or missing dependency caused a failure. Simulation therefore provides more than a substitute for deployment. It offers the experimental control needed to replay episodes, isolate individual factors, and trace long-horizon failures across perception, communication, planning, and execution.

Existing simulators support indoor interaction, driving, and open-ended language agents~\cite{savva2019habitat,li2023behavior,wang2023voyager}. Recent urban systems extend this
landscape to video cognition and embodied question answering~\cite{zhao2025urbanvideo,zhao2025cityeqa,han2026wildcity}, as well as textual or population-scale city-agent simulation~\cite{xu2023urban,yan2024opencity}.
Yet these directions typically isolate perception, a single embodiment class, or population-level behavior. Integrated support remains limited for heterogeneous agents that perceive rendered urban
scenes through their sensors and execute coordinated actions in a shared city-scale environment, with synchronized physics and episode-level traces of model-driven decisions and execution.

We present \textbf{Lingjing}, a simulation platform for heterogeneous multi-agent embodied intelligence in dynamic cities. Figure~\ref{fig:lingjing-overview} illustrates its city-scale environment and agent loop across perception, reasoning, action, and communication. Lingjing unifies three capabilities.

First, Lingjing provides a realistic, evolving urban substrate grounded in streamed geographic data, extending beyond fixed scene boundaries and coupling high-fidelity 3D environments with structured, queryable state. It represents residential, industrial, commercial, campus, logistics, and coastal districts together with dynamic traffic, weather, communication coverage, resource budgets, and mission events. The environment supports UAVs, ground robots, autonomous vehicles, fixed infrastructure, and optional wide-area sensing, while integrating AirSim~\cite{shah2017airsim}, CARLA~\cite{dosovitskiy2017carla}, MuJoCo~\cite{todorov2012mujoco}, and maritime simulation under shared physics rules.

Second, a Gym-like interface exposes user-defined ReAct~\cite{yao2022react} agents that couple observation, reasoning, and control to translate free-form language into executable actions. Natural-language missions specify roles, observation channels, communication links, and action spaces, allowing rule-based and model-based policies under star or broadcast communication to share one interaction contract.

Third, Lingjing defines complex urban embodied tasks across search, navigation, command, and coordination while making each run accountable through attribution-oriented records. Each replayable episode captures structured state, visual observations, actions, inter-agent messages, relation-graph updates, token and cost traces, and evaluation signals. These records support visual-attention inspection, reasoning and communication analysis, context-drift diagnosis, cost and communication accounting, relation-level metrics, and causal interventions on recurring failure patterns.

Across nine urban tasks covering navigation, inspection, delivery, emergency response, and replanning, we evaluate five commercial and seven open-source vision-language backbones. We compare backbone performance, communication modes, and costs, trace subtask error propagation, and test scalability and robustness across workload, sensing, and appearance variations to assess model capability, coordination efficiency, and failure modes.

This work makes three contributions.
\begin{itemize}
    \item We introduce Lingjing, a dynamic multi-engine urban platform with a common interface for heterogeneous embodied agents, structured communication, and engine-based adjudication, providing a controlled testbed for studying and evaluating coordination across representative urban missions.
    \item We provide a framework for user-defined urban embodied tasks, instantiated with nine navigation, inspection, delivery, emergency-response, and replanning tasks, with replayable traces for attributing failures across perception, localization, handoff, coordination, and execution.
    \item We conduct extensive experiments with twelve commercial and open-weight backbones, revealing current performance gaps and providing practical guidance for improving urban embodied agents.
\end{itemize}

\newcolumntype{L}[1]{>{\raggedright\arraybackslash}m{#1}}
\newcolumntype{C}[1]{>{\centering\arraybackslash}m{#1}}

\section{Related Work}
\label{sec:related}

\begin{table*}[t]
	\centering
	\scriptsize
	\setlength{\tabcolsep}{2.2pt}
	\newcommand{\ljtopcell}[2][l]{\begin{tabular}[t]{@{}#1@{}}#2\end{tabular}}
	\begin{tabular*}{\textwidth}{@{\extracolsep{\fill}} l l c l l l c c @{}}
		\toprule
		\textbf{Domain}
		& \textbf{Platforms}
		& \ljtopcell[c]{\textbf{State}\\\textbf{Evolution}}
		& \ljtopcell{\textbf{State}\\\textbf{Representation}}
		& \ljtopcell{\textbf{Agent}\\\textbf{Types}}
		& \textbf{Communication}
		& \ljtopcell[c]{\textbf{Relation}\\\textbf{Evidence}}
		& \ljtopcell[c]{\textbf{Failure}\\\textbf{Attribution}} \\
		\midrule
		
		Indoor
		& \ljtopcell{Habitat, iGibson}
		& Static/local
		& \ljtopcell{RGB-D, semantics,\\objects}
		& Indoor agents
		& None
		& State logs
		& None \\
		
		Physics
		& \ljtopcell{OmniGibson, Isaac Sim,\\Genesis}
		& Static/local
		& \ljtopcell{Poses, physics\\states}
		& Robots
		& Task-specific
		& State logs
		& Basic \\
		
		Driving
		& CARLA, MetaDrive
		& Domain-specific
		& \ljtopcell{RGB-D/LiDAR,\\road/traffic}
		& \ljtopcell{Vehicles,\\pedestrians}
		& Environment-mediated
		& Event traces
		& Basic \\
		
		Minecraft
		& MineDojo, Voyager
		& Static/local
		& Game, inventory
		& Minecraft agents
		& None
		& State logs
		& None \\
		
		Social
		& \ljtopcell{Generative Agents,\\MindCraft, Project Sid}
		& Domain-specific
		& \ljtopcell{Dialogue, memory,\\context}
		& \ljtopcell{Language/social\\agents}
		& \ljtopcell{Peer-to-peer\\dialogue}
		& Event traces
		& Basic \\
		
		Open-world
		& SimWorld
		& Mission-coupled
		& \ljtopcell{Observations,\\scene graph}
		& \ljtopcell{Humans, robots,\\vehicles}
		& Broadcast/shared
		& Event traces
		& Basic \\
		
		\ljtopcell{Urban\\embodied}
		& \ljtopcell{MetaUrban, EmbodiedCity,\\COL, CARLA-Air,\\UrbanVideo-Bench, CityEQA}
		& Domain/mission
		& \ljtopcell{Rendered/video,\\GPS/map}
		& \ljtopcell{Robots, UAVs,\\vehicles}
		& Environment-mediated
		& Frames/events
		& Basic \\
		
		\ljtopcell{Urban\\population}
		& \ljtopcell{UGI, OpenCity, CitySim,\\MobileCity, CAMS}
		& Daily/mobility
		& \ljtopcell{Text, schedules,\\trajectories}
		& Human/LLM agents
		& Not central
		& Aggregate traces
		& None \\
		
		\midrule
		
		\ljtopcell{\textbf{Urban}\\\textbf{Multi-agent}}
		& \textbf{Lingjing (Ours)}
		& \textbf{Mission-coupled}
		& \textbf{\ljtopcell{RGB-D,\\city/task, relation}}
		& \textbf{\ljtopcell{UAVs/UGVs,vehicles,\\ships}}
		& \textbf{Star/Broadcast}
		& \textbf{Typed relations}
		& \textbf{End-to-end} \\
		
		\bottomrule
	\end{tabular*}
		\caption{Comparison of Lingjing with representative embodied simulators. State evolution denotes temporal scope rather than physics fidelity: static/local, domain-specific, or mission-coupled. Relation evidence comprises state logs, event traces, or typed task relations (information handoff, cooperation, competition, guidance, and dependency). Failure attribution is absent (None), diagnostic only (Basic), or links outcomes to agents, actions, messages, and task relations (End-to-end). Environment-mediated communication uses shared world state.}
	\label{tab:platform_comparison}
\end{table*}

\paragraph{Embodied and Urban Simulation.} Table~\ref{tab:platform_comparison} situates Lingjing among representative systems. Indoor and physics platforms support navigation and interaction~\cite{savva2019habitat,shen2020igibson,li2023behavior,gao2026nvidia,genesisdocs}, while driving and outdoor simulators extend embodied evaluation to traffic and city settings~\cite{dosovitskiy2017carla,li2022metadrive,metaurban,gao2024embodiedcity}. Open-ended environments add language-guided skills and social behavior in game or Unreal worlds~\cite{fan2022minedojo,wang2023voyager,park2023generative,white2025collaborating,al2024project,ren2025simworld}. UrbanVideo-Bench evaluates urban Video-LLM cognition~\cite{zhao2025urbanvideo}. CityEQA studies active embodied question answering through a Planner, Manager, and Actor hierarchy~\cite{zhao2025cityeqa}, while Sentinel and AirCopBench address cooperative spatial reasoning and multi-drone perception~\cite{lin2026sentinel,zha2026aircopbench}. These systems advance urban perception but usually focus on one embodiment or task family. Platforms for population simulation pursue a different goal. UGI links CityGPT agents to a textual city simulator and knowledge graph~\cite{xu2023urban}. OpenCity scales daily activity simulation to $10{,}000$ agents~\cite{yan2024opencity}. CitySim models individual and aggregate behavior with tens of thousands of agents~\cite{bougie2025citysim}, while MobileCity evaluates multimodal mobility over $4{,}000$ agents~\cite{ye2025mobilecity}. CAMS uses geospatially informed CityGPT agents to generate mobility trajectories~\cite{du2025cams}. These platforms emphasize scalable behavioral simulation rather than rendered closed-loop action. A broader taxonomy appears in~\cite{li2025embodied}. Lingjing instead provides a shared rendered city for heterogeneous embodied agents, enabling coordinated and traceable physical execution.

\paragraph{Long-Horizon Embodied Task Benchmarks.} Long-horizon tasks require agents to maintain goals, track intermediate states and dependencies, and recover across extended action sequences. ALFRED~\cite{shridhar2020alfred} and TEACh~\cite{padmakumar2022teach} study multi-step household instruction following and dialog-guided task completion. CALVIN~\cite{mees2022calvin}, RoboCerebra~\cite{han2026robocerebra}, and VLABench~\cite{zhang2025vlabench} evaluate language-conditioned manipulation, high-level reasoning, and long-horizon execution, while BEHAVIOR and BEHAVIOR-1K~\cite{srivastava2022behavior,li2023behavior} cover diverse activities with realistic physical interactions. Most benchmarks, however, retain single-agent execution in bounded environments with predefined tasks, limiting evaluation of heterogeneous role organization, cross-agent dependencies, joint-plan adaptation under evolving conditions, and failure diagnosis across agents and decisions.
\section{The Lingjing Platform}
\label{sec:platform}
\begin{figure*}[ht]
	\centering
	\includegraphics[width=0.85\textwidth]{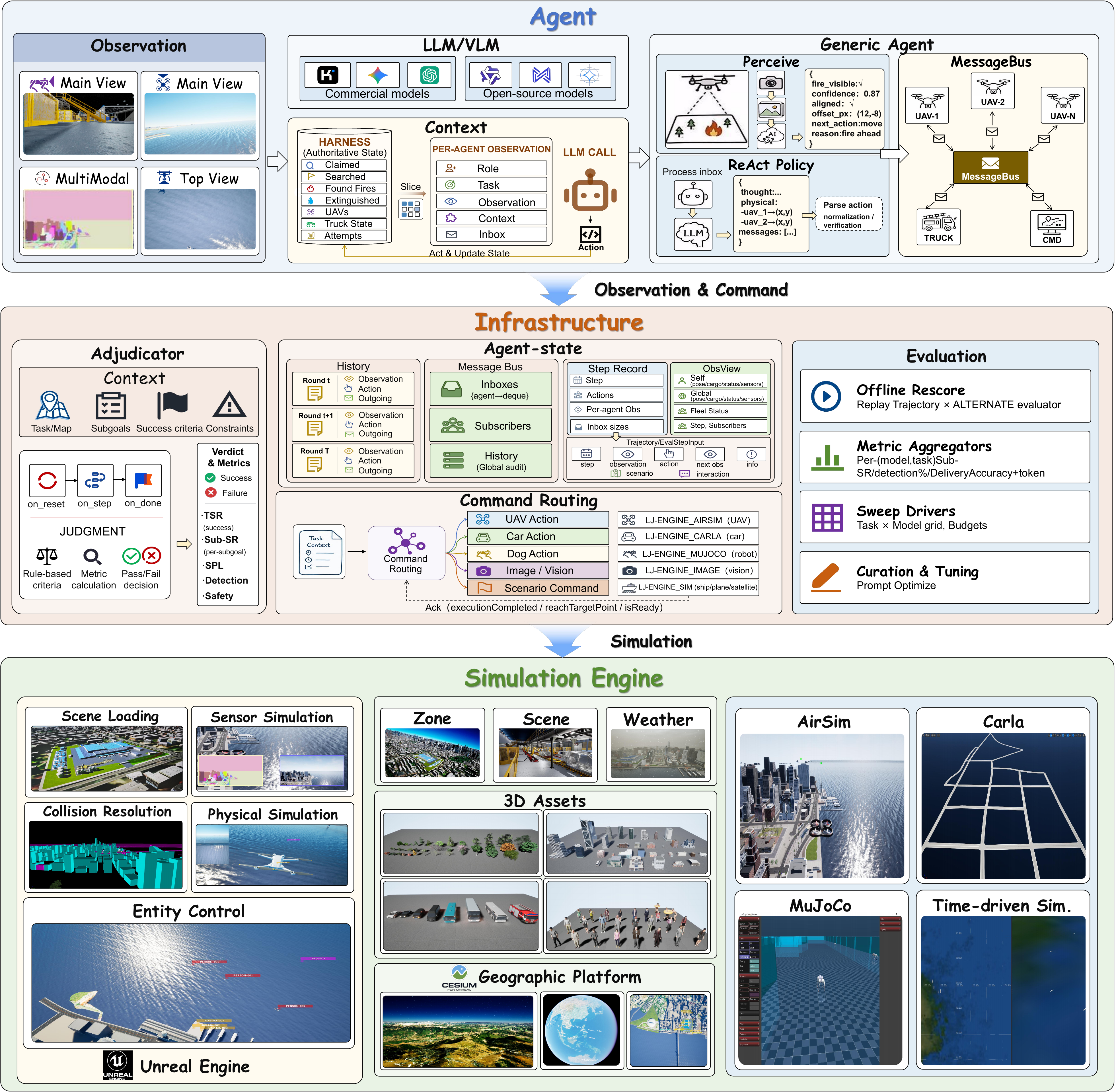}
	\caption{Lingjing's three-layer architecture: an urban physics engine, shared infrastructure, and heterogeneous ReAct agents.}
	\label{fig:framework}
\end{figure*}

Lingjing comprises an urban physics engine, shared infrastructure, and a framework for heterogeneous embodied agents, supporting mission execution and evaluation (Fig.~\ref{fig:framework}).

\paragraph{Urban Physics Engine.}
Urban missions require heterogeneous agents to act in one evolving city, yet domain-specific simulators model only fragments of this setting. To our knowledge, Lingjing is the first city-scale platform to couple heterogeneous embodied agents across multiple specialized physics engines within a single rendered execution loop. It anchors AirSim~\cite{shah2017airsim}, CARLA~\cite{dosovitskiy2017carla}, and MuJoCo~\cite{todorov2012mujoco} to a common Cesium geodetic frame~\cite{cesiumunreal} with Unreal rendering, then maps their observations, commands, and state transitions into a shared spatiotemporal state at every kernel tick. Changes produced by one backend therefore become part of the world observed by all agents. This cross-engine closed loop combines realistic urban geometry with evolving traffic, weather, and mission conditions, while exposing rendered observations (e.g., RGB, depth, and semantic views) alongside structured road graphs and city state for reproducible evaluation.

\begin{algorithm}[t]
	\caption{Kernel-simulation step.}
	\label{alg:kernel}
	\begin{algorithmic}[1]
		\Require actions/messages $(U,M)$; $W=(B,S,G,\mathcal{E},R)$
		\If{reset is requested} \Comment{scene control}
		\State dispatch $S$ and await all engines' \textsc{ready}
		\State $(U,M)\gets(\emptyset,\emptyset)$
		\EndIf
		\State $I_a\gets\emptyset$ for each active agent $a$
		\ForAll{$m_a\in M$} \Comment{communication}
		\State validate and route $m_a$ via $B$ \Comment{coordinator/all peers}
		\EndFor
		\ForAll{$u_a\in U$} \Comment{command routing}
		\State validate and dispatch $u_a$ to its engine
		\EndFor
		\State advance one tick and collect $O$, receipts, and $T$
		\State $G\gets\Call{Update}{G,T,O,\text{receipts}}$
		\State $\mathcal{V}\gets\Call{Adjudicate}{G,T,S}$
		\State $(r,d)\gets\Call{Evaluate}{\mathcal{E},G,\mathcal{V}}$
		\State $R.\Call{Log}{G,T,O,I,U,M,\mathcal{V},r,d}$
		\State \Return $O,I,\mathcal{V},r,d$
	\end{algorithmic}
\end{algorithm}

\paragraph{Infrastructure}
\label{sec:plat:rec}
The shared infrastructure turns multi-engine execution into engine-grounded, replayable evidence. The kernel manages scenario transitions and routes validated actions and messages through a common bus (Algorithm~\ref{alg:kernel}). Its adjudicator derives events and outcomes directly from synchronized engine state, while pluggable evaluators use them for online termination. A task-indexed recorder aligns each decision with its observations, messages, relation updates, and resulting world transition. This provenance enables subtask and communication analysis as well as failure attribution in the following experiments.

\paragraph{Agent Framework}
\label{sec:plat:agents}
The agent framework defines a policy-independent contract between embodiment and coordination. Through a Gym-like reset/step interface, scenarios specify capabilities, resource limits, roles, and communication relations (Appendix~A.1 and A.2). Each ReAct agent connects a dynamically configurable MLLM to a common perception, reasoning, and action loop. The model grounds visual and structured observations, reasons over task commitments, and issues physical actions or messages, providing the basis for embodied task execution and the shared scaffold for MLLM evaluation (\S\ref{sec:experiments}). Because roles and communication relations remain separate from embodiment, the same asset can act alone or join a UAV and UGV team without changing its interface. The contract supports both rule and model policies under \emph{star} or \emph{broadcast} coordination (\S\ref{sec:exp:comm}).

\paragraph{Runtime Loop}
The runtime closes the perception-action loop through deterministic ticks that place every engine and agent on one execution timeline. At tick $t$, the kernel executes $(U_t,M_t)$, advances the engines, adjudicates events, and records the transition. The agent runtime consumes $(O_t,I_t)$ and returns $(U_{t+1},M_{t+1})$, with pending requests retained across ticks. In the appendix, Algorithm~1 formalizes this decision-to-outcome boundary for the default runtime and task orchestrators.

\section{Urban Missions and Agent Coordination}
\label{sec:vla}
\begin{figure*}[t]
	\centering
	\includegraphics[width=0.9\textwidth,height=0.32\textheight,keepaspectratio]{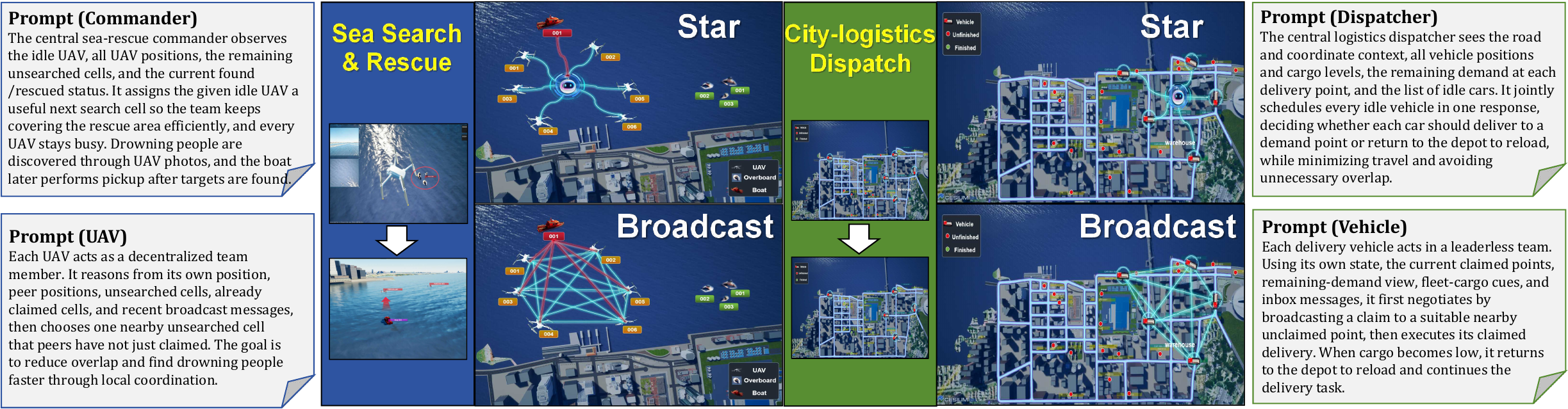}
	\caption{Representative star and broadcast coordination in sea rescue and city-logistics dispatch. Star assigns work through a commander or dispatcher, whereas broadcast agents coordinate through peer messages and shared task state.}
	\label{fig:comm-scenario-examples}
\end{figure*}
Each Lingjing mission is declared in a scenario file dispatched at reset. It specifies the environment, assets and initial states, natural-language goal, high-level perceive--decide--act workflow, and coordination relations, decoupling mission design from policies and task-specific solvers. Seeded selection reproducibly varies active assets and targets while preserving the layout, enabling controlled team-size and workload sweeps. Targets remain hidden from agents, and engine state adjudicates success. Appendix~A.5 provides an example.

\paragraph{Geographic Reconstruction}
An L11 global imagery layer supports the reconstructed scenes in Ordos, Inner Mongolia, and Shanghai, with selected task regions refined to L18. BIGEMAP provides the imagery and building and road vectors.\footnote{\url{https://www.bigemap.com/}} The vectors drive ArcGIS CityEngine\footnote{\url{https://www.esri.com/en-us/arcgis/products/arcgis-cityengine}} rules that sample configurable building heights and assign facade textures. Models are exported in WGS84 and rendered with automatic LOD4 optimization. We plan to release the scenes, task templates, and seed mappings.

\paragraph{Evaluation Tasks.}
\label{sec:bench:tasks}
To evaluate embodied agents in Lingjing, we instantiate nine urban tasks (Table~\ref{tab:tasks}), spanning single-agent perception and navigation as well as heterogeneous missions involving hidden-target localization, cross-embodiment handoffs, and long-horizon coordination. UFR, SSR, and CD support both star and broadcast agent communication modes for controlled comparisons, while capability tags summarize the primary requirements of each task.

\begin{table}[t]
	\centering\scriptsize
	\setlength{\tabcolsep}{2.5pt}
	\renewcommand{\arraystretch}{1.0}
	\resizebox{0.6\columnwidth}{!}{%
		\begin{tabular}{@{}lllll@{}}
			\toprule
			\textbf{ID} & \textbf{Scenario} & \textbf{Assets} & \textbf{Coord.} & \textbf{Tags} \\
			\midrule
			VDN & Visual dog navigation       & UGV          & single      & L,N,V \\
			BI & Bridge fracture inspection & UAV          & single      & L,N,P,V \\
			LD & Last-mile delivery          & UAV+\allowbreak CAR+\allowbreak UGV   & star        & L,N,P,V,C \\
			IDF & Industrial dog firefighting & UGVs       & broadcast    & L,N,P,V,C \\
			UPR & UAV-guided pipeline repair  & UAV+\allowbreak UGV       & star        & L,N,P,V,C \\
			UFR & UAV--fire-truck response  & UAVs+\allowbreak CAR       & star, broadcast & L,N,P,V,C \\
			CR & Congestion-aware replanning & UAV+\allowbreak CAR       & star        & N,P,V,C \\
			SSR & Sea search and rescue       & UAVs+\allowbreak USV       & star, broadcast & L,N,P,V,C \\
			CD & City-logistics dispatch     & CARs          & star, broadcast & N,P,C \\
			\bottomrule
		\end{tabular}
	}
	\vspace{0.2cm}
	\caption{Nine urban tasks. CAR denotes a CARLA vehicle. L/N/P/V/C denote localization, navigation, planning, visual perception, and coordination.}
	\label{tab:tasks}
\end{table}


\paragraph{Inter-Agent Coordination Relations.}
Multi-agent scenarios use star routing through a coordinator or broadcast over a shared bus with fixed per-agent policies, enabling controlled comparisons in \S\ref{sec:exp:comm}. Scenario specifications and traces label \emph{information handoff} (evidence transfer), \emph{cooperation} (complementary work), \emph{competition} (resource contention), \emph{guidance} (directed assignment), and \emph{dependency} (prerequisite). These labels describe task structure rather than separate experimental factors (Fig.~\ref{fig:comm-scenario-examples}). The long-horizon last-mile delivery task illustrates how several relations compose. Cooperation links UAV route inspection, vehicle transport, and robot-dog final delivery. Information handoff carries UAV congestion evidence to the CARLA routing stage before vehicle execution. After the vehicle reaches the stop node, the UAV provides guidance by generating dog waypoints from its top-down view. Dependencies therefore order route inspection before vehicle execution and vehicle arrival before dog deployment. The recorded chain links route inspection, endpoint grounding, waypoint selection, legal traversal, and target recognition, exposing how upstream errors block terminal delivery in the provenance and staged-intervention analysis (Appendix~B.7).

\section{Experiments}
\label{sec:experiments}

\begin{table*}[htpb]
	\centering
	\begin{minipage}[t]{0.73\textwidth}
		\vspace{0pt}
		\centering\scriptsize
		\renewcommand{\arraystretch}{0.92}
		\setlength{\tabcolsep}{2.6pt}
		\resizebox{\linewidth}{!}{%
			\begin{tabular}{@{}l*{13}{c}@{}}
				\toprule
				\textbf{Backbone} & VDN & BI & LD & IDF & UPR & UFR$^{s}$ & UFR$^{b}$ & CR & SSR$^{s}$ & SSR$^{b}$ & CD$^{s}$ & CD$^{b}$ & Avg \\
				\midrule
				\multicolumn{14}{@{}l}{\new{\emph{Commercial models}}} \\
				GPT-5.5~\cite{gpt55}         & \new{\textbf{72}} & \new{\textbf{40}} & \new{\underline{48}} & \new{79} & \new{\underline{47}} & \new{\textbf{63}} & \new{\underline{67}} & \new{\underline{74}} & \new{\underline{79}} & \new{\underline{84}} & \new{\textbf{90}} & \new{\textbf{78}} & \new{\textbf{66}} \\
				GPT-5.4~\cite{gpt54}         & \new{\underline{66}} & \new{34} & \new{43} & \new{\textbf{86}} & \new{42} & \new{58} & \new{64} & \new{\textbf{80}} & \new{75} & \new{82} & \new{\underline{86}} & \new{\underline{74}} & \new{63} \\
				Gemini-3.1-Pro~\cite{team2023gemini}  & \new{64} & \new{\underline{37}} & \new{\textbf{52}} & \new{\underline{81}} & \new{44} & \new{\underline{61}} & \new{\textbf{70}} & \new{69} & \new{\textbf{83}} & \new{\textbf{87}} & \new{80} & \new{69} & \new{\underline{64}} \\
				Kimi-K2.5~\cite{kimik25}       & \new{51} & \new{23} & \new{36} & \new{54} & \new{28} & \new{47} & \new{41} & \new{61} & \new{46} & \new{58} & \new{60} & \new{45} & \new{45} \\
				Qwen3-VL-Flash~\cite{bai2025qwen3}  & \new{39} & \new{25} & \new{31} & \new{46} & \new{\textbf{50}} & \new{38} & \new{47} & \new{52} & \new{49} & \new{55} & \new{51} & \new{44} & \new{43} \\
				\midrule
				\multicolumn{14}{@{}l}{\new{\emph{Open-source models}}} \\
				Qwen3-VL-2B~\cite{bai2025qwen3}     & \new{38} & \new{16} & \new{27} & \new{29} & \new{18} & \new{30} & \new{27} & \new{46} & \new{35} & \new{33} & \new{34} & \new{36} & \new{30} \\
				Qwen3-VL-4B~\cite{bai2025qwen3}     & \new{44} & \new{24} & \new{25} & \new{38} & \new{26} & \new{33} & \new{41} & \new{45} & \new{40} & \new{44} & \new{47} & \new{35} & \new{36} \\
				Qwen3-VL-8B~\cite{bai2025qwen3}     & \new{48} & \new{31} & \new{37} & \new{49} & \new{29} & \new{46} & \new{50} & \new{56} & \new{45} & \new{59} & \new{55} & \new{48} & \new{45} \\
				InternVL3.5-8B~\cite{internvl35}  & \new{43} & \new{28} & \new{30} & \new{47} & \new{23} & \new{41} & \new{36} & \new{53} & \new{44} & \new{47} & \new{52} & \new{39} & \new{39} \\
				Pixtral-12B~\cite{pixtral12b}     & \new{52} & \new{21} & \new{35} & \new{40} & \new{27} & \new{34} & \new{40} & \new{47} & \new{38} & \new{50} & \new{46} & \new{42} & \new{39} \\
				MiniCPM-V-4.5~\cite{minicpmv45}   & \new{34} & \new{22} & \new{21} & \new{33} & \new{20} & \new{32} & \new{29} & \new{39} & \new{34} & \new{32} & \new{40} & \new{31} & \new{30} \\
				Gemma-3-4B~\cite{gemma3}      & \new{30} & \new{17} & \new{23} & \new{26} & \new{16} & \new{25} & \new{31} & \new{42} & \new{29} & \new{35} & \new{32} & \new{27} & \new{27} \\
				\bottomrule
			\end{tabular}
		}
		\caption{Engine-adjudicated TSR (\%) on nine tasks. Superscripts $s$ and $b$ denote star and broadcast. Avg averages modes within UFR, SSR, and CD before the nine-task mean. Bold and underlined values are the best and second-best.}
		\label{tab:main}
	\end{minipage}
	\hfill
	\begin{minipage}[t]{0.225\textwidth}
		\vspace{0pt}
		\centering\scriptsize
		\renewcommand{\arraystretch}{0.75}
		\setlength{\tabcolsep}{2pt}
		\resizebox{\linewidth}{!}{%
			\begin{tabular}{@{}llrr@{}}
				\toprule
				\textbf{Task} & \textbf{Metric} & \textbf{Star} & \textbf{Broadcast} \\
				\midrule
				Logistics (CD) & TSR & \textbf{62} & 48 \\
				& Msg & \textbf{6} & 38 \\
				& KB & \textbf{0.7} & 1.9 \\
				& Rnd & \textbf{4} & 12 \\
				& Gap & \textbf{10} & 22 \\
				& Rep & \textbf{14} & 20 \\
				\midrule
				Fire (UFR) & TSR & 55 & \textbf{60} \\
				& Msg & \textbf{6} & 61 \\
				& KB & \textbf{0.5} & 3.0 \\
				& Rnd & \textbf{5} & 9 \\
				& Gap & \textbf{13} & 15 \\
				& Rep & 17 & \textbf{11} \\
				\midrule
				Sea (SSR) & TSR & 58 & \textbf{74} \\
				& Msg & \textbf{5} & 9 \\
				& KB & \textbf{0.4} & 0.7 \\
				& Rnd & \textbf{5} & 8 \\
				& Gap & 16 & \textbf{11} \\
				& Rep & 23 & \textbf{9} \\
				\bottomrule
			\end{tabular}
		}
		\caption{Communication modes on matched instances with fixed qwen3.7-plus. Bold marks the better mode.}
		\label{tab:comm}
	\end{minipage}
\end{table*}
\subsection{Experimental Setup}
\label{sec:exp:setup}

\textbf{Backbones.} We evaluate twelve multimodal backbones in one scaffold: five commercial models (GPT-5.5~\cite{gpt55}, GPT-5.4~\cite{gpt54}, Gemini-3.1-Pro~\cite{team2023gemini}, Kimi-K2.5~\cite{kimik25}, and Qwen3-VL-Flash~\cite{bai2025qwen3}) and seven open-weight models (Qwen3-VL-2B/4B/8B~\cite{bai2025qwen3}, InternVL3.5-8B~\cite{internvl35}, MiniCPM-V-4.5~\cite{minicpmv45}, Gemma-3-4B~\cite{gemma3}, and Pixtral-12B~\cite{pixtral12b}). Per task, all share observation encoding, prompt, action schema, and communication interface without specific fine-tuning. Superscripts $s$ and $b$ denote star and broadcast. Each diagnostic states its backbone, while scaffold analyses use qwen3.7-plus~\cite{qwen37plus} to balance performance and cost.

\textbf{Metrics.} Task success rate (TSR) is the percentage of episodes satisfying all engine-adjudicated terminal conditions within budget. Adjudication uses engine proximity and events, not model self-reports, with geometry- and motion-based acceptance regions fixed across backbones and communication modes. Intermediate detections only unlock later stages. Avg is the unweighted mean TSR across tasks, and subtask completion rate (SCR) measures partial progress in multi-target or multi-stage missions. Other metrics are defined at first use. Appendices~A.3 and~B.1 give all task conditions and definitions.

%

\textbf{Implementation Details.} Open-source models use vLLM in bfloat16 with a 16K context on four NVIDIA A800 GPUs, while commercial models use provider endpoints. Within each task, all models share the prompt and $2048$-token output limit, use provider-specific near-greedy decoding, and perform both visual analysis and action selection. Unless stated otherwise, each model and diagnostic condition uses $100$ matched seeded instances from fixed templates and initial poses, with seeds also controlling entity subsets, while larger-sample studies report their sizes explicitly. Agents receive visual and structured observations, history, messages, and coarse search regions but not hidden-target coordinates, then emit physical actions and optional messages. Fixed time and step budgets bound each episode. Vision timeouts count as no detections, while infrastructure failures are tracked separately. Additional communication, intervention, appearance, component, and prompt analyses appear in Appendix~B.


\begin{figure}[htpb]
	\centering
	\begin{minipage}[t]{0.7\linewidth}
		\vspace{0pt}
		\centering
		\includegraphics[width=\linewidth]{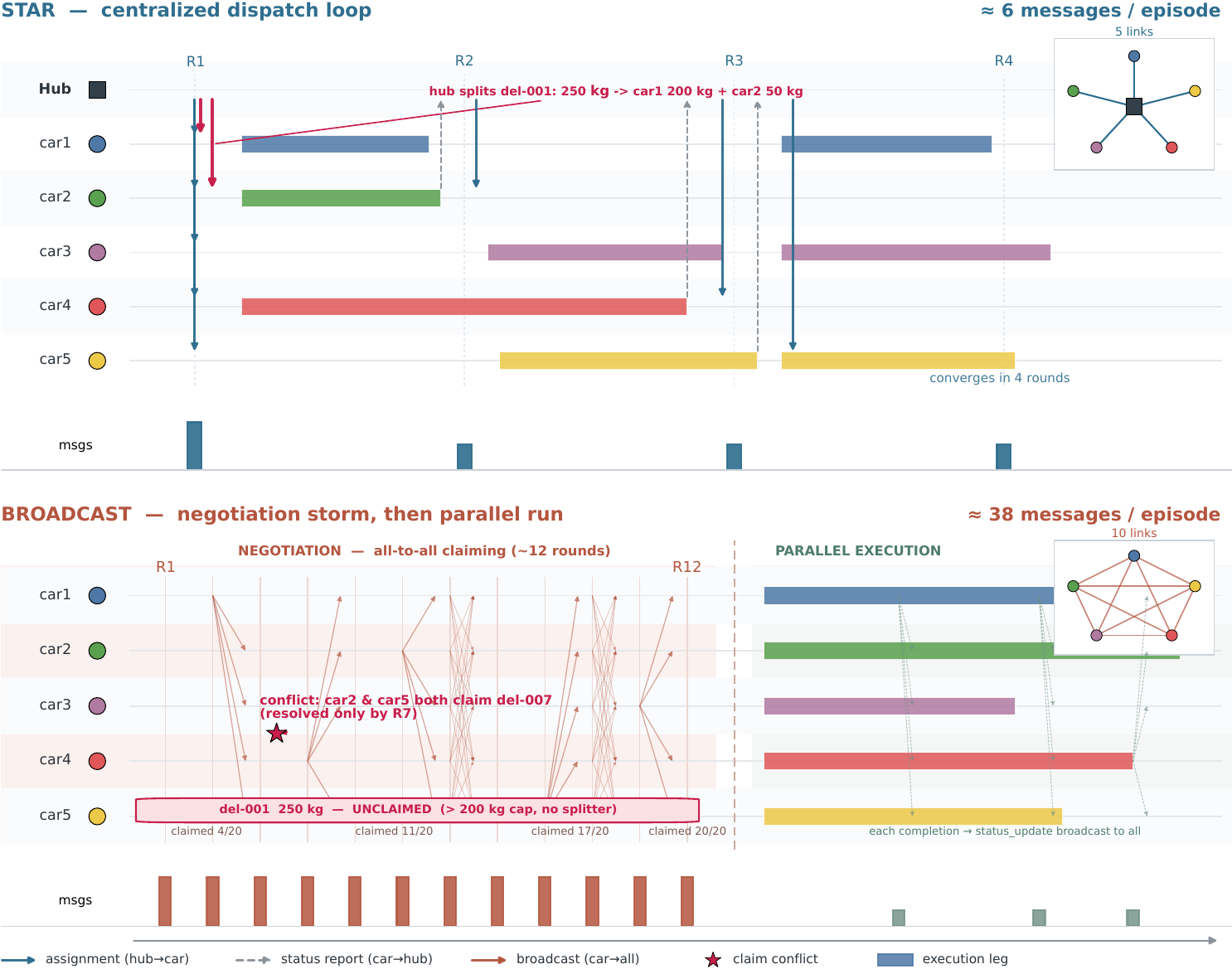}
		\\[-0.3em]\footnotesize (a) Execution process and communication traces.
	\end{minipage}
	\vspace{0.4em}
	\begin{minipage}[t]{0.7\linewidth}
		\vspace{0pt}
		\centering
		\includegraphics[width=\linewidth]{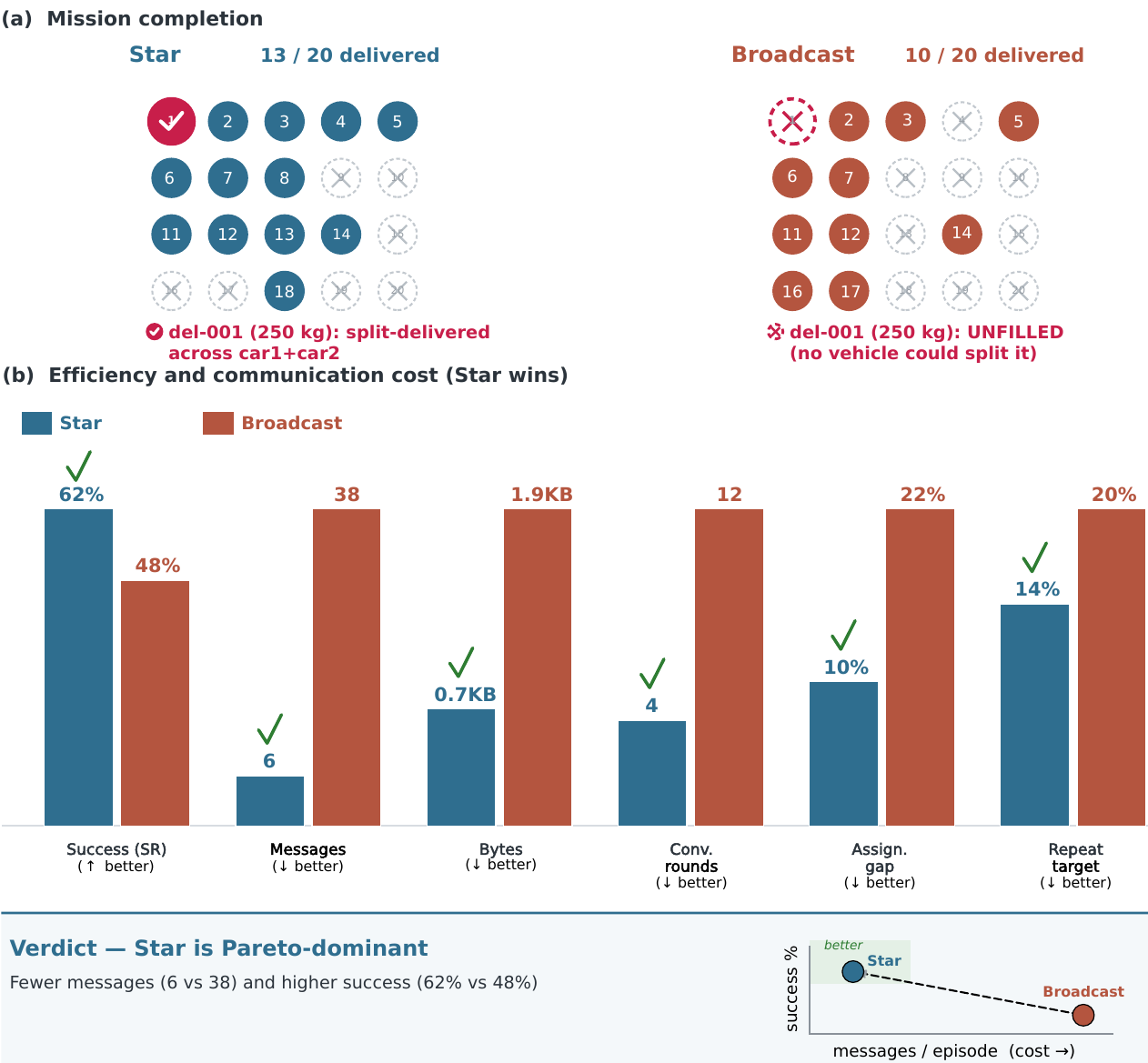}
		\\[-0.3em]\footnotesize (b) Task result analysis.
	\end{minipage}
	\caption{\new{City-logistics comparison with fixed qwen3.7-plus. Top: representative traces, where star splits the $250$\,kg order \texttt{del-001} and broadcast leaves it unassigned. Bottom: matched-instance outcomes, where star increases TSR and reduces all reported coordination measures.}}
	\label{fig:logistics-process}
\end{figure}

\subsection{Main Results}
\label{sec:exp:main}
Table~\ref{tab:main} shows GPT-5.5 leading with an Avg of $66$, followed by Gemini-3.1-Pro at $64$ and GPT-5.4 at $63$. Across the twelve task-mode columns, GPT-5.5 leads five, Gemini-3.1-Pro four, GPT-5.4 two, and Qwen3-VL-Flash one. Qwen3-VL-8B is the strongest open-source model at $45$, tying Kimi-K2.5 and exceeding its 2B and 4B variants by $15$ and $9$ points, although task-level scaling is not monotonic. BI, UPR, and LD remain hardest, with best TSRs of $40$, $50$, and $52$, exposing bottlenecks in localization, constrained planning, and multi-stage execution beyond visual recognition. Broadcast outperforms star for $8/12$ UFR and $10/12$ SSR backbones, whereas star wins $11/12$ CD comparisons. Routing gains therefore depend on task structure rather than message reach alone. Next, Section~\ref{sec:exp:comm} analyzes communication cost.

\subsection{Communication Modes and Cost}
\label{sec:exp:comm}
%

\begin{figure}[htpb]
	\centering
	\includegraphics[width=0.6\linewidth]{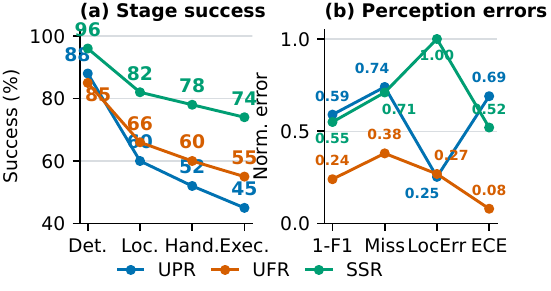}
	\caption{\new{Subtask provenance and perception diagnosis. (a) Cumulative stage success. (b) $1-\mathrm{F1}$, miss rate, normalized LocErr, and ECE, where higher is worse.}}
	\label{fig:subtask-provenance}
\end{figure}

With qwen3.7-plus and matched instances fixed, only routing varies: star uses a coordinator, whereas broadcast reaches all peers (Table~\ref{tab:comm}). TSR is task success rate, while Msg, KB, and Rnd report message count, payload in kilobytes, and convergence rounds. Assignment gap (Gap) is the percentage deviation from the reference-optimal assignment, while repeated-target rate (Rep) is the percentage of target assignments or visits that are duplicates. Lower values are better for all five coordination measures. In CD, the top panel of Fig.~\ref{fig:logistics-process} shows star splitting the $250$\,kg order while broadcast leaves it unassigned. The bottom panel shows TSR rising from $48$ to $62$, messages falling from $38$ to $6$, and the assignment gap narrowing from $22\%$ to $10\%$. Supplementary sea-rescue traces and matched-instance results appear in the appendix B.2. UFR shows a weaker trade-off: broadcast adds five TSR points but increases messages from $6$ to $61$. Thus, star favors structured allocation, whereas broadcast benefits rapid discovery sharing at higher communication cost.

\subsection{Subtask Provenance}
\label{sec:exp:subtask}
With fixed qwen3.7-plus across $100$ episodes per mission, we reconstruct cumulative provenance chains for UAV-guided pipeline repair (UPR), UAV--fire-truck response (UFR), and sea search and rescue (SSR). The recorder parses each model text output and aligns its reported target and coordinates with engine-tracked ground truth at the same timestamp. Det. requires at least one matched positive report for every target, Loc. additionally requires one reported coordinate per target within its task-specific acceptance region, Hand. a successful transfer to the downstream executor, and Exec. engine-confirmed sealing, extinguishing, or rescue. Since each stage includes all predecessors, panel (a) reports cumulative attrition. Panel (b) aggregates the same aligned calls. F1 is the harmonic mean of frame-level precision and recall, miss rate the share of all aligned frames containing a missed target, and ECE the ten-bin weighted confidence--accuracy gap. These metrics characterize detection, while LocErr characterizes localization as the median Euclidean world-coordinate error over matched true-positive reports, normalized by the largest mission value. Figure~\ref{fig:subtask-provenance} identifies Det.-to-Loc. grounding as the largest single-stage loss at $28$, $19$, and $14$ points for UPR, UFR, and SSR. The cumulative Loc.-to-Exec. losses, spanning handoff and execution, are $15$, $11$, and $8$ points. Despite per-frame miss rates of $74\%$ for UPR and $71\%$ for SSR, repeated observations raise cumulative detection to $88\%$ and $96\%$. UFR achieves the best F1 ($0.76$) and ECE ($0.08$), while UPR has the worst ECE ($0.69$). At the default SSR altitude of $100$\,m, its maximum LocErr of $190$\,m appears as $1.00$ after normalization. Overall, repeated sensing improves detection coverage, but localization remains the main bottleneck. Appendix~B.7 further analyzes staged error interventions for the last-mile delivery task.

\subsection{Scaling and Robustness}
\label{sec:exp:scaling}
\begin{figure}[htbp]
	\centering
	\begin{minipage}[t]{0.6\linewidth}
		\vspace{0pt}
		\centering
		\includegraphics[width=0.9\linewidth]{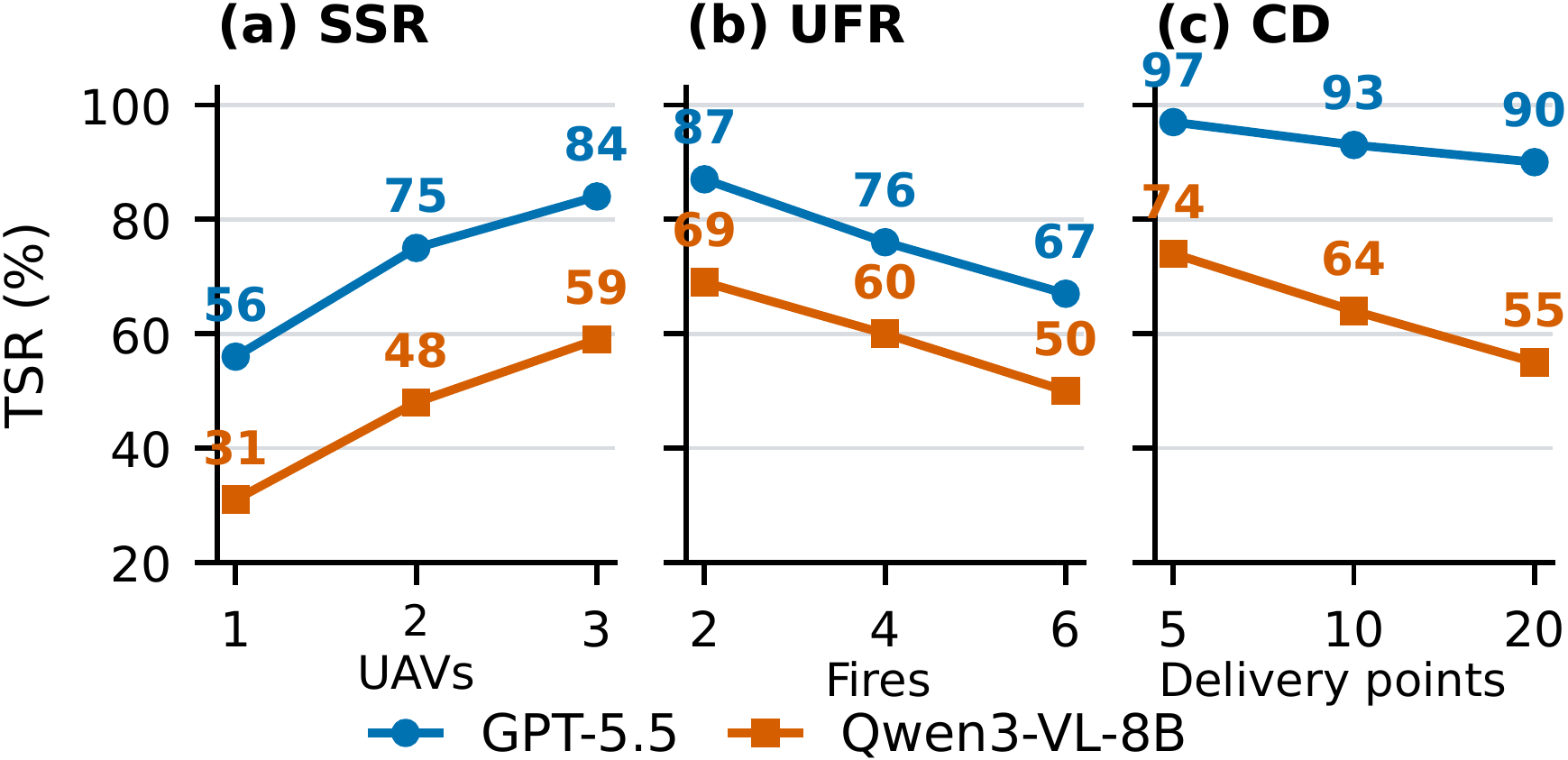}
		\\[-0.2em]\footnotesize (a) Team-size and workload sweeps.
	\end{minipage}
	\vspace{0.4em}
	\begin{minipage}[t]{0.6\linewidth}
		\vspace{0pt}
		\centering
		\includegraphics[width=0.9\linewidth]{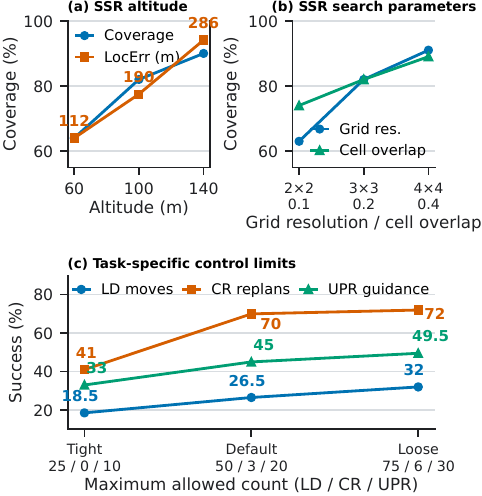}
		\\[-0.2em]\footnotesize (b) Sensing and control-budget sweeps.
	\end{minipage}
	\caption{Scaling and robustness analyses. Top: team-size and workload sweeps over SSR UAV count, UFR fire count, and CD delivery-point count. Bottom: robustness sweeps over SSR sensing settings and LD, CR, and UPR control budgets.}
	\label{fig:scale-robust}
\end{figure}


Scaling uses GPT-5.5 and Qwen3-VL-8B with $100$ matched episodes per point. Under broadcast, SSR sweeps UAVs from $1$ to $3$ with three victims, UFR fires from $2$ to $6$ with five UAVs, and CD under star delivery points from $5$ to $20$ with five vehicles. Shared seeds select targets and recompute ground truth and terminal criteria. Figure~\ref{fig:scale-robust} (top) shows diminishing capacity returns and workload degradation. Adding a second SSR UAV raises TSR by $19/17$ points and a third by $9/11$ for GPT-5.5/Qwen3-VL-8B. From lowest to highest workload, UFR falls by $20/19$ points and CD by $7/19$. GPT-5.5 leads throughout, while Qwen3-VL-8B is more sensitive to CD load.

Fixing qwen3.7-plus, we run $200$ matched episodes per point to reduce variance and sweep SSR altitude ($60/100/140$\,m), grid resolution ($2\!\times\!2/3\!\times\!3/4\!\times\!4$), overlap ($0.1/0.2/0.4$), and control budgets for LD moves ($25/50/75$), CR replans ($0/3/6$), and UPR guidance ($10/20/30$). Success is cumulative S4 legal-edge arrival for LD and engine-adjudicated destination arrival or leak sealing for CR/UPR, with other settings fixed. Figure~\ref{fig:scale-robust} (bottom) shows a coverage-precision trade-off and diminishing budget returns. Altitude $60\to140$\,m raises coverage $64\%\to90\%$ but LocErr $112\to286$\,m. Grid and overlap yield coverage $63\%\to82\%\to91\%$ and $74\%\to82\%\to89\%$, respectively. Tight-to-default budgets add $8.0/29/12$ points for LD/CR/UPR, versus $5.5/2/4.5$ from default-to-loose. Defaults capture most available gains.

\section{Conclusion}
We introduce Lingjing, a dynamic urban platform unifying multiple physics engines and heterogeneous agents with star and broadcast communication, engine-based evaluation, and replayable diagnostic traces. Across nine tasks and twelve multimodal backbones, grounding and long-horizon execution remain bottlenecks, coordination gains depend on task structure, added capacity has diminishing returns, and heavier workloads reduce success. Lingjing provides a controlled testbed for evaluating and diagnosing heterogeneous multi-agent embodied intelligence in cities.


\bibliographystyle{plain}
\bibliography{aaai2027}
\newpage
\appendix
\setcounter{table}{0}
\setcounter{figure}{0}
\setcounter{algorithm}{0}
\setcounter{listing}{0}
\renewcommand{\thetable}{R\arabic{table}}
\renewcommand{\thefigure}{R\arabic{figure}}
\renewcommand{\thelisting}{R\arabic{listing}}

\section{Lingjing Specification Details}
\label{app:bench}

\subsection{Asset Interfaces}
\label{app:assets}

Table~\ref{tab:agents} summarizes the observations and commands exposed to
policies. Assets exchange structured messages through the shared bus, and
\texttt{takePhoto} requests a camera observation. Command coordinates
$(x,y,z)$ use the shared world frame, where $x$ and $y$ are planar axes and
$z$ is elevation.

\begin{table}[htpb]
	\centering\small
	\setlength{\tabcolsep}{3.5pt}
	\begin{tabular}{@{}p{1.85cm}p{1.55cm}p{3.80cm}@{}}
		\toprule
		\textbf{Asset (engine)} & \textbf{Observations} & \textbf{Interface commands} \\
		\midrule
		UAV (AirSim) & top-down/front RGB, depth & \texttt{takeoff}, \texttt{setDestination}$(x,y,z)$, \texttt{forward}/\allowbreak\texttt{backward}/\allowbreak\texttt{left}/\allowbreak\texttt{right}/\allowbreak\texttt{up}/\allowbreak\texttt{down}, \texttt{stop}, \texttt{takePhoto} \\
		Autonomous vehicle (CARLA) & front RGB, telemetry & \texttt{setDestination}$(x,y)$, \texttt{forward}/\allowbreak\texttt{left}/\allowbreak\texttt{right}/\allowbreak\texttt{stop} \\
		Quadruped robot (MuJoCo) & front RGB & \texttt{forward}/\allowbreak\texttt{left}/\allowbreak\texttt{right}/\allowbreak\texttt{stop}, \texttt{takePhoto} \\
		Surface vessel (time-driven) & position telemetry & \texttt{moveTo}$(x,y)$ \\
		\bottomrule
	\end{tabular}
	\caption{Supported assets, observations, and policy-facing commands.}
	\label{tab:agents}
\end{table}

\subsection{Task Profiles and Illustrations}
\label{app:bench:taskprofiles}

\definecolor{ljMorandiPaper}{HTML}{F3F0EA}
\definecolor{ljMorandiPanel}{HTML}{FBFAF7}
\definecolor{ljMorandiBorder}{HTML}{C9C0B3}
\definecolor{ljMorandiPrompt}{HTML}{EFE8DE}
\definecolor{ljMorandiPromptBorder}{HTML}{B8AA9A}
\definecolor{ljSage}{HTML}{8E9A8B}
\definecolor{ljSlate}{HTML}{73808A}
\definecolor{ljClay}{HTML}{B18F7A}
\definecolor{ljDust}{HTML}{A89192}
\definecolor{ljMist}{HTML}{829A9A}
\definecolor{ljOlive}{HTML}{9B9A7B}

\newcommand{\ljtaskprofile}[5]{%
	\par\smallskip
	\noindent
	\begingroup
	\setlength{\fboxsep}{0pt}%
	\setlength{\fboxrule}{0.45pt}%
	\fcolorbox{ljMorandiBorder}{ljMorandiPanel}{%
		\begin{minipage}{\dimexpr\linewidth-2\fboxrule\relax}
			\colorbox{#2}{%
				\begin{minipage}{\dimexpr\linewidth-2\fboxrule\relax}
					\vspace{2pt}
					\hspace*{0.5em}\begin{minipage}{\dimexpr\linewidth-1.0em\relax}
						\raggedright\footnotesize\textcolor{white}{\textbf{#1}}
					\end{minipage}
					\vspace{2pt}
			\end{minipage}}%
			\vspace{0.45em}
			\hspace*{0.55em}\begin{minipage}{\dimexpr\linewidth-1.1em\relax}
				\footnotesize
				\textbf{Task introduction.} #3\par\smallskip
				\textbf{Prompt.}\par\vspace{0.2em}
				\begingroup
				\setlength{\fboxsep}{4pt}%
				\fcolorbox{ljMorandiPromptBorder}{ljMorandiPrompt}{%
					\begin{minipage}{\dimexpr\linewidth-2\fboxsep-2\fboxrule\relax}
						\ttfamily\scriptsize\raggedright #4
				\end{minipage}}%
				\endgroup
				\par\smallskip
				\textbf{Simulation-engine view.}\par\vspace{0.2em}
				\begingroup
				\setlength{\fboxsep}{0pt}%
				\fcolorbox{ljMorandiPromptBorder}{ljMorandiPaper}{%
					\begin{minipage}[c][0.77\linewidth][c]{\dimexpr\linewidth-2\fboxrule\relax}
						\centering
						\if\relax\detokenize{#5}\relax
						\scriptsize\textcolor{ljSlate}{Corresponding simulator view.}
						\else
						\includegraphics[width=\linewidth,height=0.75\linewidth,keepaspectratio]{#5}
						\fi
				\end{minipage}}%
				\endgroup
			\end{minipage}
			\vspace{0.55em}
	\end{minipage}}%
	\endgroup
	\par\smallskip
}

\ljtaskprofile{VDN: Visual dog navigation}{ljSage}{%
	The robot dog starts outside a residential community and must enter the
	compound, search from its egocentric front camera, and reach the target
	white building where pedestrians stand or walk in front of the facade. The
	episode tests whether one ground agent can use a reference image and the
	current front view together: the policy must match facade color, height,
	pedestrian context, and local layout while avoiding large blind moves. The
	action space contains \texttt{forward}, \texttt{left}, \texttt{right}, and
	\texttt{stop}. Forward commands are clipped to a short step and turns are kept
	within a bounded angle, so success depends on repeated visual grounding rather
	than a single long jump.}{%
	You are the navigation policy for one robot dog in a residential community.
	Inputs are the target-building reference image, the current front-view image,
	pose history, and recent actions. First compare the reference and current view:
	main facade color, height, windows, entrances, pedestrian presence, sidewalk
	objects, and nearby landmarks. Then estimate whether the target is left,
	right, ahead, behind, or not visible, and note obstacles that constrain motion.
	Return exactly one command from forward(distance 5 to 25 m), left(degrees 1 to 90),
	right(degrees 1 to 90), or stop. Use short forward moves when uncertain. The
	engine registers success when the dog enters the configured target region. Use
	\texttt{stop} once the matched facade is close, centered, and visually
	consistent with the reference.}{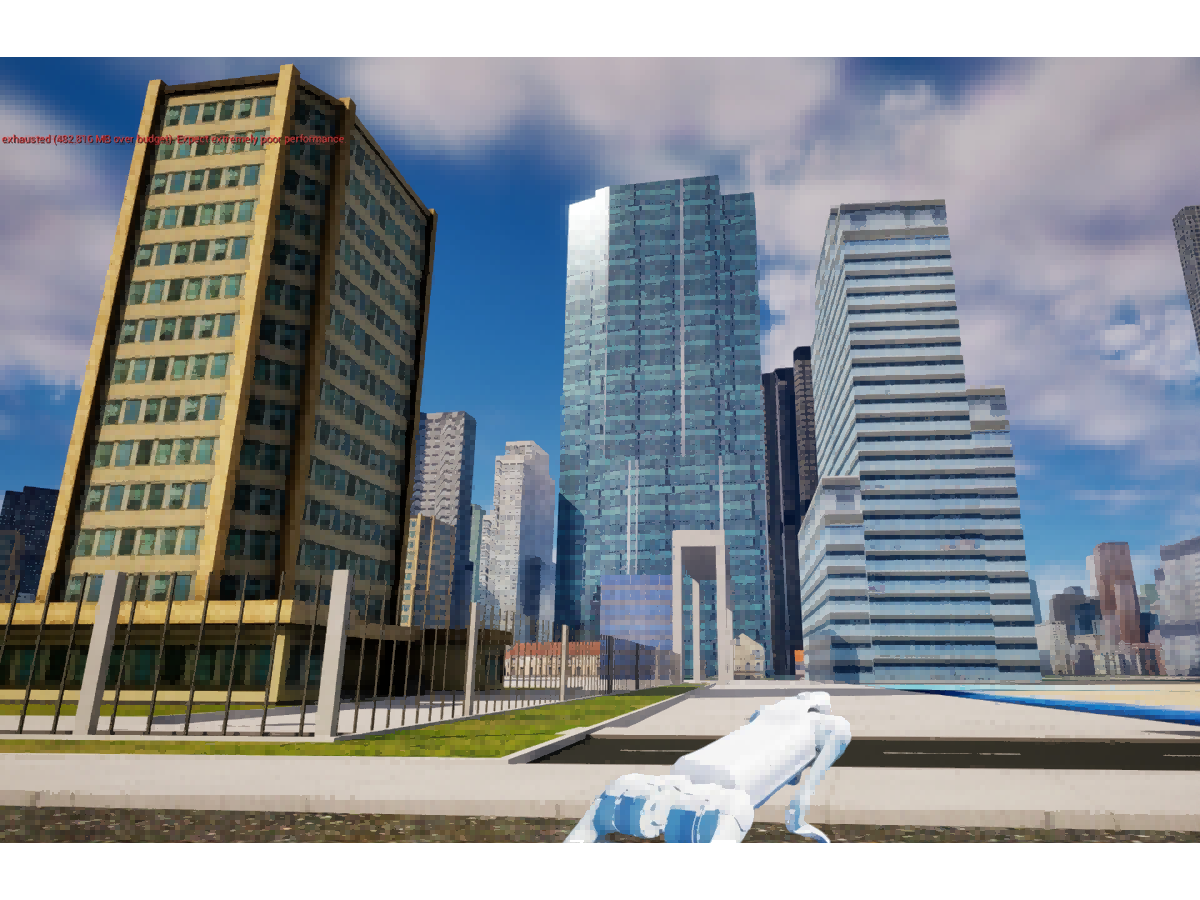}

\ljtaskprofile{LD: Last-mile delivery}{ljSlate}{%
	A UAV, autonomous vehicle, and robot dog complete a staged delivery. CARLA
	first returns a candidate vehicle route. The UAV inspects its waypoints with
	top-down images and reports congestion for replanning. The vehicle executes
	only a verified route to the stop node. The dog is then deployed, follows
	waypoints generated from the UAV view, and uses its front camera to confirm the
	target house. The task couples route inspection, evidence handoff, vehicle
	transport, aerial guidance, and final visual navigation.}{%
	You coordinate a UAV, delivery vehicle, and robot dog. Inputs include the
	candidate route, UAV route images, blockage history, vehicle state, dog pose,
	and target-house observations. Inspect route waypoints in order. If congestion
	is detected, report the blocked waypoint and request replanning. Otherwise,
	verify the route before vehicle execution. After vehicle arrival, generate dog
	waypoints from the UAV view and use the dog's front image to confirm the target
	house. Return the current stage, evidence, and next executable command. Declare
	completion only after engine-confirmed final delivery.}{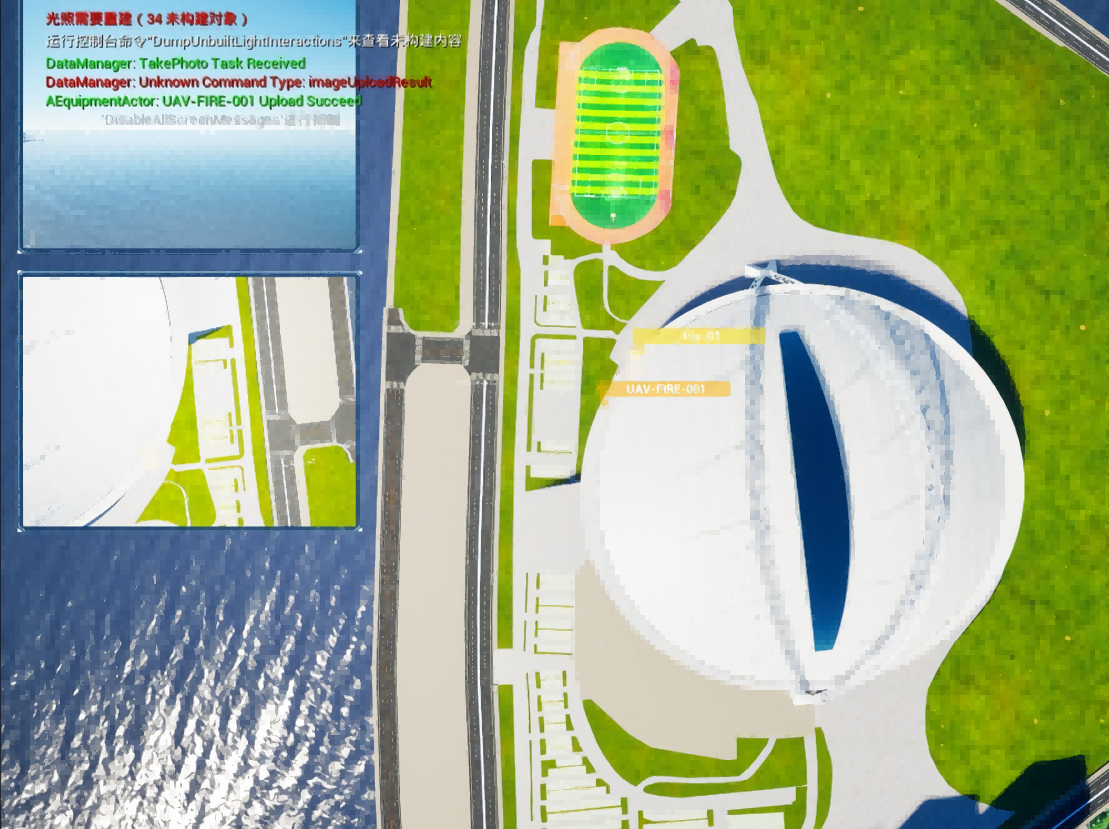}

\ljtaskprofile{BI: Bridge fracture inspection}{ljClay}{%
	The UAV climbs to an inspection altitude, follows the bridge deck from land
	toward open water, and checks each span for a fracture, interruption, or
	collapse. The bridge may leave the top-down frame as the UAV advances, so the
	policy must recover the structure from front-view evidence instead of drifting
	inland or stopping on an intact section. The objective is complete coverage of
	the bridge until the damaged span is found. Success requires an
	engine-confirmed inspection within the fractured region.}{%
	You are a UAV inspector following a bridge from land toward open water. Inputs
	are the current top-down image, optional front-view image, pose, previous span
	status, and inspection direction. Identify the bridge deck axis, piers, deck
	edges, lane markings, shadows, and water boundary. For each step, decide
	whether the current span is intact, partially interrupted, collapsed, or out of
	view. Continue along the bridge when spans are intact. If the deck leaves the
	top-down frame, use the front view to turn back toward the bridge or sea side,
	not inland. Report fracture\_found only when a discontinuity or collapsed span
	is visually clear. Otherwise return continue\_inspection.}{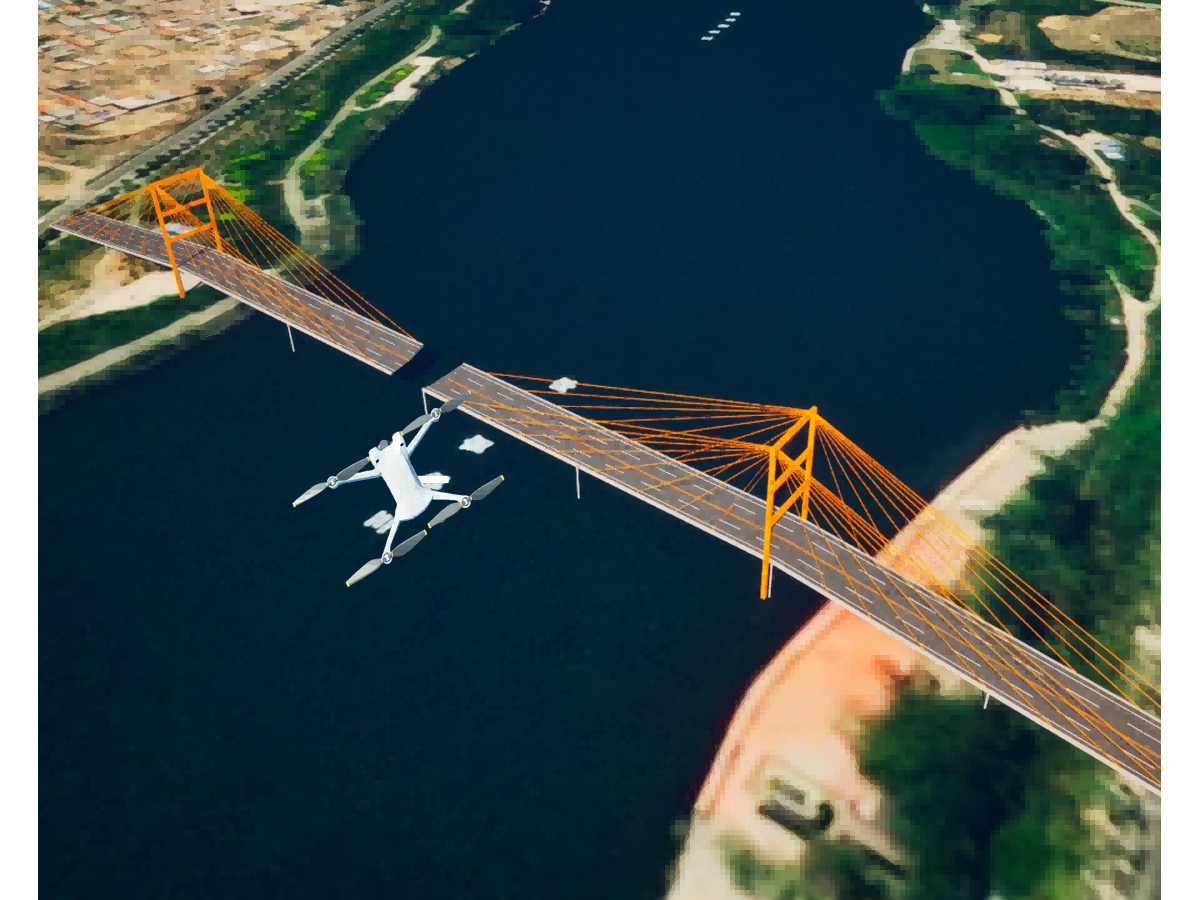}

\ljtaskprofile{UFR: UAV and fire-truck response}{ljDust}{%
	Five UAVs patrol assigned urban search areas to detect hidden building fires,
	while one fire truck acts as the shared extinguishing resource. The agents do
	not receive the true fire coordinates. Each UAV captures top-down
	images, uses the vision model to determine whether a fire exists in its area,
	and shares the candidate location through the configured communication mode. The truck
	is then routed to confirmed fires one by one, and the episode ends only after
	the simulator emits extinguishing events for all fires. This task measures
	multi-UAV area assignment, hidden-target detection, dispatch arbitration, and
	contention for a single ground resource.}{%
	You are the coordinator for a multi-UAV and single-fire-truck response team.
	Inputs are UAV search areas, current UAV images, fire-detection messages,
	estimated coordinates, truck status, extinguished-fire records, and outstanding
	targets. For each UAV, decide whether it should continue patrol, refine a
	detection, or report a confirmed fire. Merge duplicate detections by area and
	coordinate, reject reports without smoke or flame evidence, and maintain the set
	of unresolved fires. Dispatch the only fire truck to one confirmed unresolved
	fire at a time, prioritizing confidence, distance, and whether another fire is
	already being handled. Return UAV assignments, selected truck target, evidence
	summary, and completion status.}{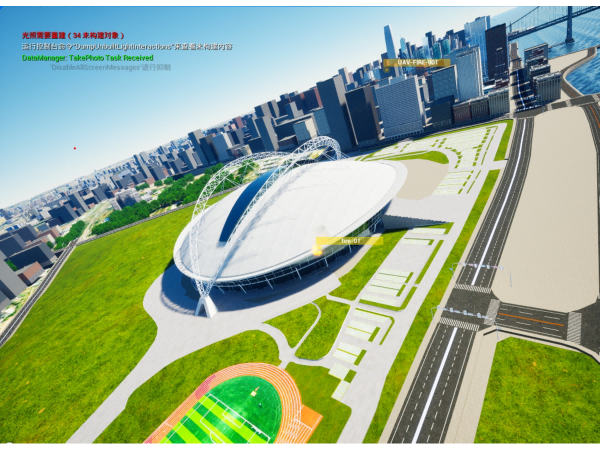}

\ljtaskprofile{SSR: Sea search and rescue}{ljMist}{%
	Multiple UAVs search a large sea region for people in the water, while one
	lifeboat or unmanned surface vessel performs physical rescue. UAVs select
	unsearched grid cells, capture top-down images, identify valid human targets,
	and hover above detections. A found-person message transfers the UAV coordinate
	to the rescue boat, which navigates to each reported location and completes the
	rescue according to engine events. The difficulty is not only perception but
	also shared state: agents must avoid duplicate cell searches, preserve the list
	of found and rescued people, and schedule a single boat under many possible
	reports.}{%
	You are the coordinator for a sea search-and-rescue team with several UAVs and
	one rescue boat. Inputs are the search grid, UAV positions, top-down images,
	previously searched cells, found-person reports, boat position, assigned
	targets, and engine rescue events. Allocate UAVs to uncovered cells to maximize
	coverage and avoid duplicate searches. In each UAV image, look for human bodies,
	life vests, small floating objects, or wake patterns. Reject waves or debris
	unless human evidence is clear. When a person is detected, send a report with
	coordinate, cell id, confidence, and whether the UAV should hover. Dispatch the
	single boat to one unresolved valid target at a time and update searched,
	found, assigned, and rescued sets after every event.}{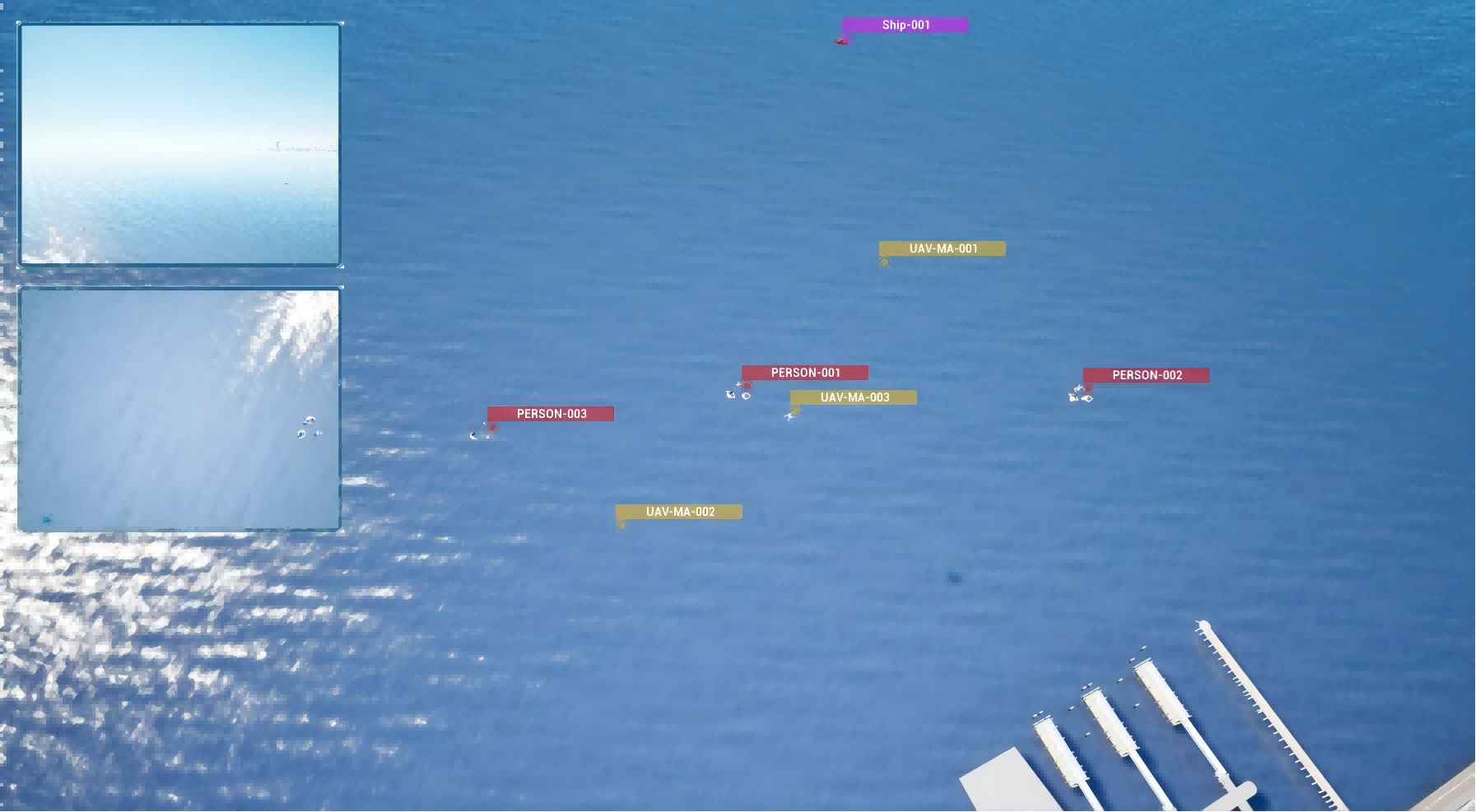}

\ljtaskprofile{CD: City-logistics dispatch}{ljOlive}{%
	Five delivery vehicles leave a warehouse to serve twenty delivery points under
	a 200 kg payload limit per vehicle. Most orders require 100 kg, but one large
	order requires 250 kg and therefore cannot be completed by a single loaded
	vehicle in one trip. Delivered quantities are updated after each route
	execution and replans until every demand is satisfied. The task asks the policy
	to reason over capacity, route order, reloads, split deliveries, and the
	communication needed to keep vehicles from duplicating work or starving a large
	order.}{%
	You are the planner for a five-vehicle urban delivery fleet. Inputs are vehicle
	locations, remaining payload, warehouse access, road-cost estimates, completed
	stops, and residual demand at twenty delivery points. Each vehicle can carry at
	most 200 kg. Build feasible routes that never exceed capacity, avoid assigning
	the same demand twice, and prefer compact routes that reduce travel and idle
	time. For normal 100 kg orders, assign one loaded visit when possible. For the
	250 kg order, explicitly split the demand across multiple vehicles or trips and
	record the remaining quantity after each execution. Return per-vehicle route,
	loaded weight, serviced demand, reload decision, and replan trigger until all
	points are complete.}{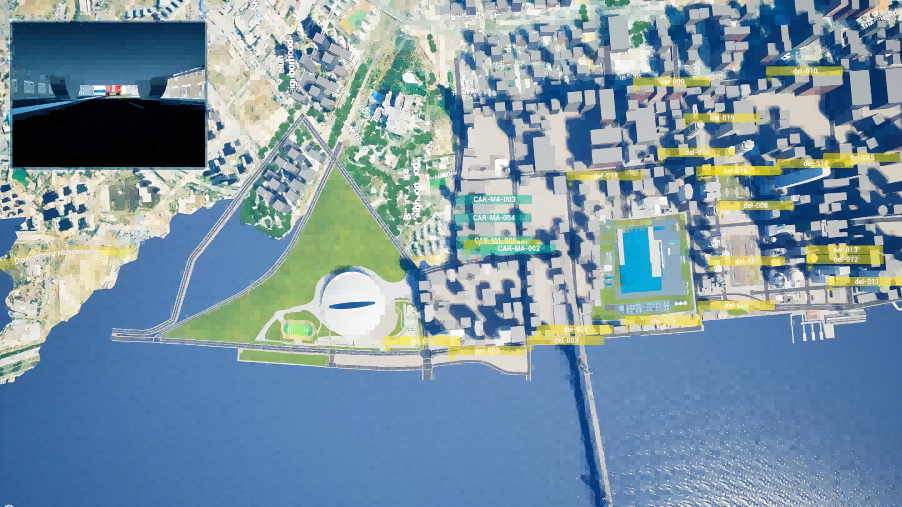}

\ljtaskprofile{IDF: Industrial dog firefighting}{ljSlate}{%
	Three Go2 robot dogs start from different entrances inside a closed industrial
	factory area. Each dog repeatedly requests a front-view image, uses the vision
	policy to detect flames, smoke, or glowing hot regions, and then executes one
	short action: turn left, turn right, move forward by at most 10 meters, or
	stop. Fire-source coordinates are hidden from the dogs and are used only by the
	engine for rendering and adjudication. When a dog gets close enough to a fire
	source, the simulator emits the extinguishing judgment. The task evaluates
	front-view fire grounding, decentralized search, short-horizon motion control,
	and progress sharing among multiple ground agents.}{%
	You are a robot-dog fire-search controller in an industrial factory. Inputs are
	the dog front-view image, current dog state, known searchable factory region,
	recent actions, and engine adjudication events. Decide whether active fire is
	visible from flame, smoke, red-hot regions, or glare. If no fire is visible,
	prefer left or right rotations to scan a new direction. Use a short forward step
	only after scanning. If fire is visible but off-center, turn toward it. Move
	forward only when the fire is roughly centered and reduce the step near the
	source. Stop only when the dog is right next to the fire and wait for the
	engine extinguishing report.}{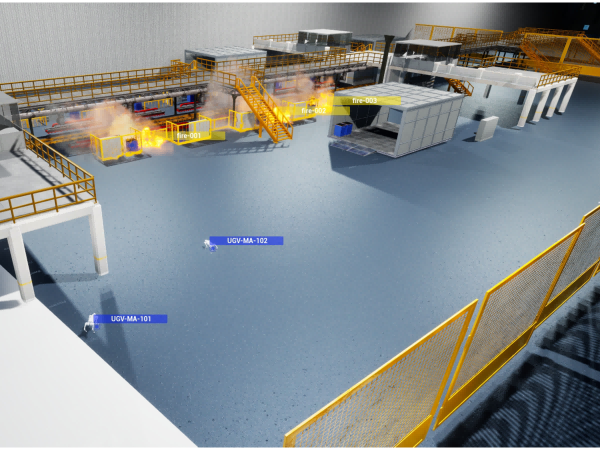}

\ljtaskprofile{UPR: UAV-guided pipeline repair}{ljClay}{%
	A UAV and a repair dog operate in an outdoor industrial pipeline area. The UAV
	takes off, flies to the known pipeline-area center, and searches only inside the
	provided boundary. The true leak coordinate is hidden. The UAV must infer
	spraying water, mist, or wet stains from top-down or oblique imagery, adjusting
	altitude within a safe band when the evidence is too coarse or the field of
	view is too narrow. After a leak is found, the UAV keeps both the leak and the
	repair dog in view and issues the dog one-step commands until the engine
	confirms sealing. The task tests aerial localization, altitude-aware inspection,
	and cross-agent guidance where the ground dog executes but does not perceive.}{%
	You are the UAV controller for industrial pipeline-leak inspection and repair.
	Inputs are the UAV top-down image, current altitude, safe altitude band,
	pipeline search boundary, repair-dog pose when available, and engine repair
	events. First search the pipeline area for spraying water, mist, or wet ground.
	Use down to inspect suspected evidence more closely and up to recover wider
	coverage, never leaving the safe altitude band or the search boundary. When a
	leak is confidently visible, switch to guidance: keep the dog visible, turn or
	move it toward the leak with forward, left, or right commands, and declare success
	only after the engine reports that the leak is sealed.}{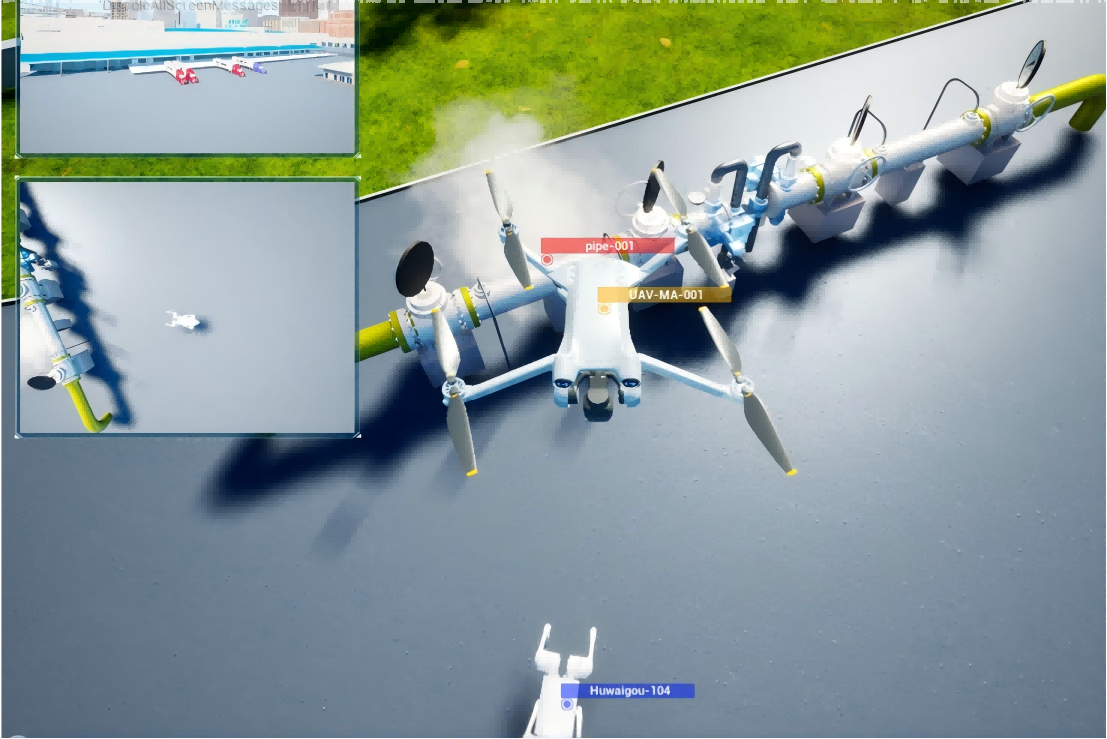}

\ljtaskprofile{CR: Congestion-aware replanning}{ljMist}{%
	A route is requested for the autonomous vehicle, and CARLA returns its
	waypoints. A UAV visits each waypoint in order and checks a top-down image for
	traffic congestion. On detection, it reports the exact waypoint via
	\texttt{trafficJam}. CARLA replans, and the UAV reinspects the complete new
	route. The vehicle remains stationary until every waypoint on one route is
	verified clear, then follows that route to the destination.}{%
	You coordinate UAV route inspection and vehicle replanning. Inputs are the
	vehicle start and destination, CARLA-reported waypoints, UAV top-down images,
	replanning history, and vehicle state. Inspect all waypoints in order. If a
	waypoint is congested, send \texttt{trafficJam} with its reported coordinates,
	request a new route, and restart inspection. Keep the vehicle stationary until
	one complete route is clear, then issue the move command. Declare success only
	when the engine confirms destination arrival.}{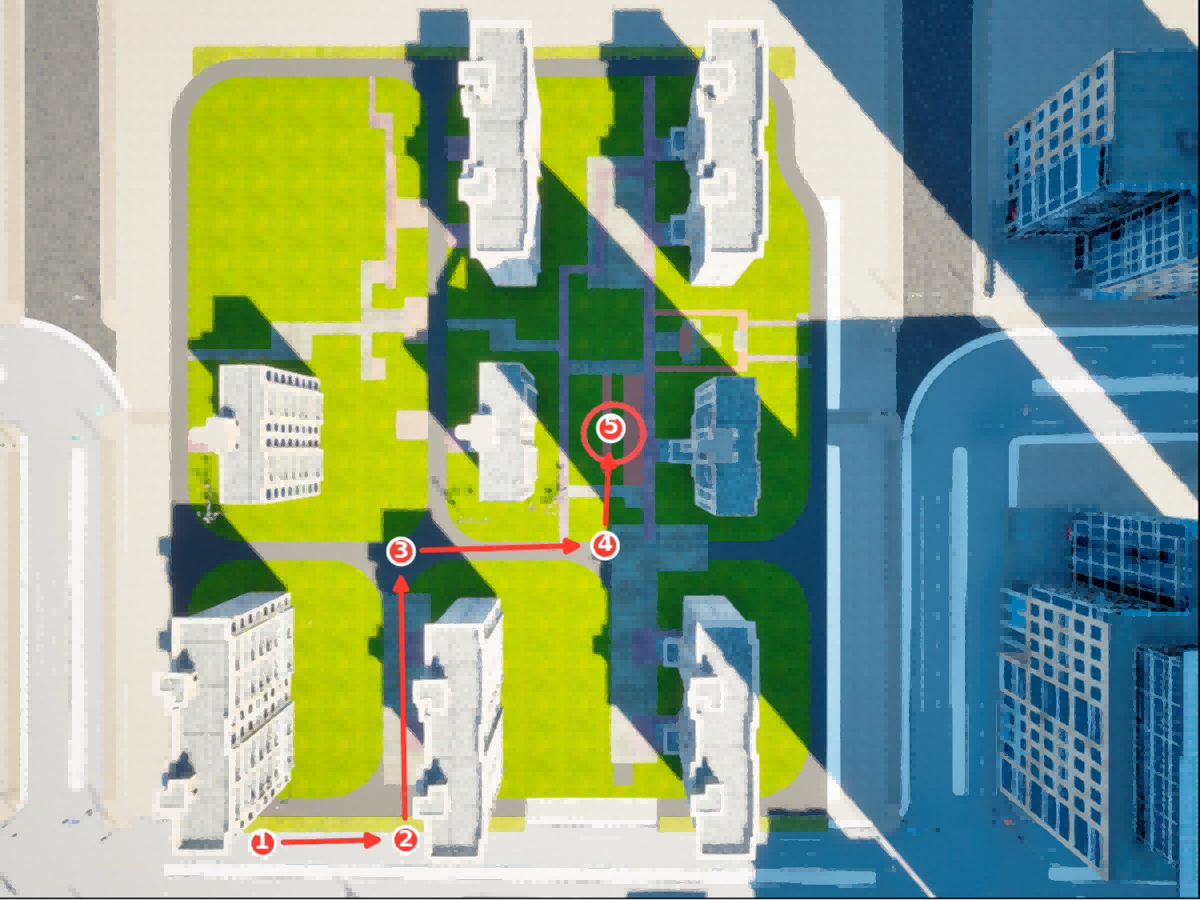}

\subsection{Task Success Adjudication}
\label{app:adjudication}

\new{TSR is the simulator's binary episode outcome. Let $d_{xy}(a,b)$ denote Euclidean distance between planar world positions $a$ and $b$. An entity argument denotes its engine-reported $XY$ position. All thresholds are inclusive. For BI, UPR, UFR, and SSR, detection (Det.), localization (Loc.), handoff (Hand.), and execution (Exec.) are recorded as separate events. Episode success requires every task terminal condition within the budget.}

\new{For visual detection adjudication, each positive model report contains a target class and a normalized image box $b=(x_{\min},y_{\min},x_{\max},y_{\max})$, where $0\le x_{\min}<x_{\max}\le1$ and $0\le y_{\min}<y_{\max}\le1$. At the same capture timestamp, the engine provides the visible target's ground-truth box $b^*$. The intersection over union $\operatorname{IoU}(b,b^*)$ is the intersection area divided by the union area. A same-class report is a true positive when $\operatorname{IoU}(b,b^*)\ge0.5$. An unmatched report is a false positive, and an unmatched visible target is a false negative. A target with no visible pixels is not evaluated in that frame. The Det. stage succeeds when every required target has at least one true-positive report during the episode. The Loc. stage additionally requires a reported world position $\widehat p$ inside the task-specific acceptance region of the engine ground truth $p^*$.}

\new{The benchmark uses task-specific inclusive radii. Extinguishing and leak handling use $2$ to $5$\,m. Aerial localization of bridge damage, fire, and congestion uses $25$\,m. Visual navigation, road-vehicle response, and delivery use $30$\,m. Vessel rescue and warehouse entry use $50$\,m. LD uses the target-house region in each scenario instance. Table~\ref{tab:task-adjudication} gives the resulting checks.}

\begin{algorithm*}[htpb]
	\caption{Operational expansion of the kernel step and enclosing agent loop.}
	\label{alg:step}
	\begin{algorithmic}[1]\scriptsize
		\Require agents $\mathcal{A}$ and policies $\{\pi_a\}$, world $W=(B,S,G,\mathcal{E},R)$ ($B$ is the message bus, $S$ is the reset specification, $G$ is the shared state registry, $\mathcal{E}$ is the evaluators, and $R$ is the recorder), limits $K$. For more notation explanations, see Table~\ref{tab:runtime-symbols}.
		\State $t\gets0$
		\Statex \textit{Reset branch of Algorithm~1}
		\ForAll{$q\in\mathcal{N}_{\mathrm{eng}}$ \textbf{in parallel}}
		\State dispatch $\Call{Reset}{S}$ to $q$
		\EndFor
		\State \textbf{reset barrier:} for at most $K_{\mathrm{engine}}$, await one $\Call{Ready}{0}$ and initial state from every $q$
		\If{the reset barrier expires or any engine rejects $S$}
		\State $R.\Call{Log}{\mathrm{infrastructure\ abort},S,\mathrm{partial\ replies}}$
		\State \Return partial trace with reason \textsc{infrastructure-abort}
		\EndIf
		\State $U_0,M_0,\widehat U_0,\widehat M_0\gets\emptyset$ and $I_0\gets\{I_0^a=\emptyset:a\in\mathcal{A}\}$
		\State $\mathcal{C}_0^M,\mathcal{C}_0^U\gets\emptyset$
		\State $(O_0^{\mathrm{raw}},\mathcal{C}_0^E,T_0)\gets\Call{AssembleBoundary}{0}$
		\State $\mathcal{C}_0\gets\mathcal{C}_0^M\cup\mathcal{C}_0^U\cup\mathcal{C}_0^E$
		\State $G\gets\Call{Update}{G,T_0,O_0^{\mathrm{raw}},\mathcal{C}_0}$
		\State $O_0\gets\Call{ProjectVisible}{G,O_0^{\mathrm{raw}},T_0,S}$
		\State $\mathcal{V}_0\gets\Call{Adjudicate}{G,T_0,S}$
		\State $(r_0,d_0)\gets\Call{Evaluate}{\mathcal{E},G,\mathcal{V}_0}$
		\State $R.\Call{Log}{0,G,T_0,O_0,I_0,U_0,M_0,\widehat U_0,\widehat M_0,\mathcal{C}_0,\mathcal{V}_0,r_0,d_0}$
		\State publish boundary output $(O_0,I_0,\mathcal{V}_0,r_0,d_0)$
		\While{$t<K_{\mathrm{tick}}$ and not $d_t$}
		\Statex \textit{Agent-runtime phase following the main-paper Runtime Loop}
		\State freeze $(O_t,I_t)$ and start a policy barrier for every active $a\in\mathcal{A}$
		\ForAll{$a\in\mathcal{A}$ \textbf{in parallel}}
		\State $(u_{t+1}^a,m_{t+1}^a)\gets\pi_a.\Call{Act}{o_t^a,I_t^a}$
		\EndFor
		\State \textbf{policy barrier:} await all calls for at most $K_{\mathrm{model}}$
		\State replace each missing or malformed return by (no-op, $\emptyset$), log the reason, and discard any later return for tick $t$
		\State $U_{t+1}\gets\{u_{t+1}^a:a\in\mathcal{A}\}$ and $M_{t+1}\gets\{m_{t+1}^a:a\in\mathcal{A}\}$ \Comment{main-paper pending inputs}
		\Statex \textit{Kernel-simulation phase expanding the communication and return steps of Algorithm~1}
		\State $I_{t+1}^a\gets\emptyset$ for every $a\in\mathcal{A}$ \Comment{main-paper inbox initialization}
		\State $(\widehat M_{t+1},\mathcal{C}_{t+1}^M)\gets\Call{ValidateAndExpand}{M_{t+1},S,t+1}$
		\State $(\widehat U_{t+1},\mathcal{C}_{t+1}^U)\gets\Call{ValidateAndResolve}{U_{t+1},G,S,t+1}$
		\State $B.\Call{Stage}{\widehat M_{t+1},t+1}$
		\State dispatch $\widehat U_{t+1}$ and request $\Call{AdvanceTo}{t+1}$ from every engine
		\State \textbf{engine barrier:} for at most $K_{\mathrm{engine}}$, await one coherent boundary state from every engine
		\State within the same barrier, await an admission receipt for every command in $\widehat U_{t+1}$
		\If{every engine supplies state but an admission receipt is missing}
		\State mark the missing action receipt as \textsc{timed-out} for boundary $t+1$
		\State retain any previously admitted long-running request
		\EndIf
		\If{an engine disconnects or does not supply a coherent boundary state}
		\State $R.\Call{Log}{\mathrm{infrastructure\ abort},S,t+1}$
		\State \Return partial trace with reason \textsc{infrastructure-abort}
		\EndIf
		\State $B.\Call{Commit}{\widehat M_{t+1},t+1}$ and $t\gets t+1$
		\State $(O_t^{\mathrm{raw}},\mathcal{C}_t^E,T_t)\gets\Call{AssembleBoundary}{t}$
		\State $\mathcal{C}_t\gets\mathcal{C}_t^M\cup\mathcal{C}_t^U\cup\mathcal{C}_t^E$ \Comment{main-paper receipts}
		\State $I_t^a\gets\Call{DeliveredTo}{a,\widehat M_t}\cup\Call{VisibleReceiptsFor}{a,\mathcal{C}_t}$ for every $a$
		\State $I_t\gets\{I_t^a:a\in\mathcal{A}\}$ \Comment{aggregate $I$ logged and returned in Algorithm~1}
		\State $G\gets\Call{Update}{G,T_t,O_t^{\mathrm{raw}},\mathcal{C}_t}$
		\State $O_t\gets\Call{ProjectVisible}{G,O_t^{\mathrm{raw}},T_t,S}$
		\State $\mathcal{V}_t\gets\Call{Adjudicate}{G,T_t,S}$
		\State $(r_t,d_t)\gets\Call{Evaluate}{\mathcal{E},G,\mathcal{V}_t}$
		\State $R.\Call{Log}{t,G,T_t,O_t,I_t,U_t,M_t,\widehat U_t,\widehat M_t,\mathcal{C}_t,\mathcal{V}_t,r_t,d_t}$
		\State publish boundary output $(O_t,I_t,\mathcal{V}_t,r_t,d_t)$
		\EndWhile
		\State \Return $R$'s trace and $\Call{TerminationReason}{\mathcal{E},G,t,K,d_t}$
	\end{algorithmic}
\end{algorithm*}

\begin{table*}[htpb]
	\centering\small
	\setlength{\tabcolsep}{4pt}
	\caption{\new{Engine adjudication conditions used to compute TSR. Distances compare model-reported or engine-reported planar positions as specified and include the stated boundary. Task IDs follow Table~2 in main text.}}
	\label{tab:task-adjudication}
	\begin{tabular}{@{}l p{0.17\textwidth} p{0.28\textwidth} p{0.39\textwidth}@{}}
		\toprule
		\textbf{ID} & \textbf{Adjudicated object} & \textbf{Engine check} & \textbf{Successful episode condition} \\
		\midrule
		VDN & Target building & $d_{xy}(\mathrm{Dog},\mathrm{Target}) \le 30\,\mathrm{m}$ & The robot dog enters the target region and the engine reports target arrival. \\
		BI & Broken-bridge location & A positive report must satisfy the box-matching rule. Its reported coordinate is accepted when $d_{xy}(\widehat p,\mathrm{BrokenBridge}) \le 25\,\mathrm{m}$. & The UAV reports the damage with a matched box and accepted coordinate, and the engine confirms successful inspection. \\
		LD & Final delivery address & The delivery vehicle completes its verified route, then the engine accepts the robot dog inside the instance-configured target-house region. & Route inspection, vehicle handoff, and final robot-dog delivery all complete in order. \\
		IDF & Factory fire & $d_{xy}(\mathrm{Dog},\mathrm{Fire}) \le 2\,\mathrm{m}$, followed by an extinguishing event & Every active fire produces an engine-confirmed extinguishing event. \\
		UPR & Pipeline leak & A positive leak report must satisfy the box-matching rule, and its coordinate must lie within the instance-configured leak-localization region before UAV guidance. Dog proximity with $d_{xy}(\mathrm{Dog},\mathrm{Leak}) \le 5\,\mathrm{m}$ triggers sealing. & Every active leak receives a matched report and accepted coordinate, and the engine confirms sealing after the repair dog enters its acceptance region. \\
		UFR & Fire location & A positive report must satisfy the box-matching rule. A coordinate satisfying $d_{xy}(\widehat p,\mathrm{Fire}) \le 25\,\mathrm{m}$ unlocks dispatch. Vehicle proximity with $d_{xy}(\mathrm{CAR},\mathrm{Fire}) \le 30\,\mathrm{m}$ triggers extinguishing. & Every active fire receives a matched report and accepted coordinate, and the dispatched vehicle extinguishes it. \\
		CR & Congestion point and destination & $d_{xy}(\mathrm{UAV},\mathrm{Congestion}) \le 25\,\mathrm{m}$ records congestion, then $d_{xy}(\mathrm{CAR},\mathrm{Destination}) \le 30\,\mathrm{m}$ triggers arrival. & The vehicle reaches its destination region after the required congestion inspection and any resulting replanning. \\
		SSR & Water victim & A positive report must satisfy the box-matching rule. A coordinate inside the configured rescue region unlocks dispatch. Vessel proximity with $d_{xy}(\mathrm{Ship},\mathrm{Victim}) \le 50\,\mathrm{m}$ triggers rescue. & Every active victim receives a matched report and accepted coordinate and produces an engine-confirmed vessel rescue event. \\
		CD & Delivery point and warehouse & $d_{xy}(\mathrm{CAR}_i,\mathrm{Delivery}_j) \le 30\,\mathrm{m}$. Here $i$ indexes vehicles and $j$ indexes delivery points. If restocking is triggered, $d_{xy}(\mathrm{CAR}_i,\mathrm{Warehouse}) \le 50\,\mathrm{m}$. & Every active delivery point has zero remaining demand, and every triggered restocking event completes. \\
		\bottomrule
	\end{tabular}
\end{table*}

\begin{listing*}[htpb]
	\begin{lstlisting}[basicstyle=\scriptsize\ttfamily,numbers=none,
		breaklines=true,columns=fullflexible,xleftmargin=0pt]
		{
			"commandType": "resetScenario",
			"taskId": "multiagentstasks_test_001",
			"scenario": {
				"commandType": "resetScenario",
				"sceneName": "UAV tracks vessel for person rescue",
				"collaborationType": "Multi-UAV image collection",
				"sceneRegion": "Multi-UAV initial area",
				"equipmentList": {
					"droneEntityList": [
					{"equipmentCode": "UAV-MA-001", "name": "drone1",
						"data": {"X": 30660.0, "Y": 62020.0, "Z": 112.0}, "raw": 0.0,
						"sensorType": "RGB/TopDown/MultiAgentPhoto"},
					{"equipmentCode": "UAV-MA-002", "name": "drone2",
						"data": {"X": 31000.0, "Y": 59010.0, "Z": 112.0}, "raw": 0.0,
						"sensorType": "RGB/TopDown/MultiAgentPhoto"},
					{"equipmentCode": "UAV-MA-003", "name": "drone3",
						"data": {"X": 27870.0, "Y": 59010.0, "Z": 112.0}, "raw": 0.0,
						"sensorType": "RGB/TopDown/MultiAgentPhoto"}
					],
					"unmannedDogEntityList": [], "autoVehicleEntityList": [],
					"satelliteEntityList": [],
					"shipEntityList": [
					{"equipmentCode": "Ship-001", "name": "Lifeboat", "side": "red",
						"data": {"X": 1240.0, "Y": 135180.0, "Z": -350.0},
						"heading": 30.0}
					],
					"planeEntityList": []
				},
				"taskMatrix": [
				{
					"taskLevel": "System",
					"task_id": "MULTIAGENTS_UAV_PHOTO_COLLECTION_001",
					"goal": "UAV tracks vessel for person rescue",
					"initial_state": {
						"weather": "Clear", "uav_count": 3,
						"rescuePersonPosition": [
						{"personId": "PERSON-001", "X": 31920.0,
							"Y": 96320.0, "Z": -350.0}
						],
						"rescueRange": [
						{"rangeId": "RANGE-001", "X": 66510.0, "Y": 108110.0, "Z": -350.0},
						{"rangeId": "RANGE-002", "X": -12370.0, "Y": 110050.0, "Z": -350.0},
						{"rangeId": "RANGE-003", "X": -12370.0, "Y": 75020.0, "Z": -350.0},
						{"rangeId": "RANGE-004", "X": 66510.0, "Y": 75490.0, "Z": -350.0}
						],
						"goalFirePosition": []
					}
				}
				]
			}
		}
	\end{lstlisting}
	\caption{Original sea-rescue scenario object. Descriptive strings are
		translated into English. Structure, identifiers, and coordinates match
		the source file.}
	\label{lst:ssr-scenario}
\end{listing*}

\subsection{Engine and Episode Protocol}
\label{app:engine-protocol}

Table~\ref{tab:engine-protocol} summarizes the runtime configuration used by
the benchmark. Simulation engines run at their native rates, while policies
interact through synchronized decision ticks. Each retained boundary records
engine frame identifiers and simulation time. Requested camera observations carry a
capture timestamp. Engine versions and scenario configurations are recorded
for each run.

\begin{table*}[htpb]
	\centering\small
	\setlength{\tabcolsep}{3pt}
	\caption{Engine, sensing, action, budget, and timeout settings. Step caps count synchronized decision ticks.}
	\label{tab:engine-protocol}
	\begin{tabular}{@{}p{0.115\textwidth}p{0.135\textwidth}p{0.13\textwidth}p{0.225\textwidth}p{0.155\textwidth}p{0.175\textwidth}@{}}
		\toprule
		\textbf{Component} & \textbf{Engine or runtime} & \textbf{Frame and decision rate} & \textbf{Sensors and policy action space} & \textbf{Default episode budget} & \textbf{Timeout handling} \\
		\midrule
		UAV & Unreal Engine~5.x rendering with AirSim control & Native physics and render rates. One synchronized snapshot per decision tick. Images are captured on request. & $1920{\times}1080$ nadir RGB in multi-agent tasks. Front RGB at the engine resolution. Depth and semantic views when declared. \texttt{takeoff}, \texttt{setDestination}, \texttt{forward}, \texttt{backward}, \texttt{left}, \texttt{right}, \texttt{up}, \texttt{down}, \texttt{stop}, \texttt{takePhoto}. & Listed in the final row & An image or analysis timeout yields ``no detection.'' \\
		Road vehicle & CARLA & Native simulation rate. One synchronized state per decision tick & Front RGB, pose, speed, route, and execution telemetry. \texttt{setDestination}, \texttt{forward}, \texttt{left}, \texttt{right}, \texttt{stop}. & Listed in the final row & A motion request remains pending until completion, rejection, or task timeout. \\
		Quadruped & MuJoCo control with Unreal Engine~5.x rendering & Native dynamics and render rates. One synchronized state per decision tick & Front RGB and pose telemetry. \texttt{forward}, \texttt{left}, \texttt{right}, \texttt{stop}, \texttt{takePhoto}. & Listed in the final row & An action timeout yields a no-op. An image timeout yields no observation. \\
		Surface vessel & Time-driven vessel simulator & Native integration rate. One synchronized state per decision tick & Position, heading, assignment, and rescue-event telemetry. \texttt{moveTo}$(x,y)$. & Listed in the final row & A rescue request remains pending until completion or task timeout. \\
		Episode control & Shared runtime & One policy call per active agent per synchronized tick & Immutable observation and inbox pair $(O_t,I_t)$. At most one physical command and a finite typed-message list per policy call. & VDN: 140 total and 40 per subtask. BI: 60. LD: 50. IDF and UPR: 300. UFR, SSR, CD, and CR: 60 & Model call: 90\,s. Missing or malformed output is replaced and logged. Reset failure, disconnect, or an incoherent snapshot is an infrastructure abort. \\
		\bottomrule
	\end{tabular}
\end{table*}

\paragraph{Failure accounting.}
A model or vision timeout yields the substitution specified in
Algorithm~\ref{alg:step}, either a no-op with an empty message list or ``no
detection.'' An invalid action or resource conflict yields a rejection receipt.
Recoverable episodes remain in the model denominator. TSR is zero when the
terminal conditions are incomplete at the tick or task limit. Valid partial
progress remains in the subtask completion rate (SCR), and admitted messages
remain in the communication metrics. Reset or synchronization failures are
infrastructure failures. They are excluded from model metrics and rerun with
the same instance and seed.

\subsection{Declarative Scenario Example}
\label{app:scenario-example}
Listing~\ref{lst:ssr-scenario} reproduces the original sea-rescue scenario file. The object structure, identifiers, and numeric values follow the source file.

\subsection{Synchronized Runtime Contract}
\label{app:runtime-step}

Algorithm~\ref{alg:step} in the appendix extends the kernel step presented in Algorithm~1 in the main text, and detailed notations are explained here (Table~\ref{tab:runtime-symbols}). It uses the same world decomposition $W=(B,S,G,\mathcal{E},R)$, where $B$ is the message bus, $S$ the reset specification, $G$ the shared state registry, $\mathcal{E}$ the evaluators, and $R$ the recorder. Let $\mathcal{A}$ be the active-agent set and let
$a\in\mathcal{A}$ index an agent with policy $\pi_a$. Let $t\ge0$ be a
committed boundary index. At boundary $t$,
$O_t=\{o_t^a\}_{a\in\mathcal{A}}$ and
$I_t=\{I_t^a\}_{a\in\mathcal{A}}$ are the policy-visible observations and
inboxes. The candidate actions $U_t=\{u_t^a\}_{a\in\mathcal{A}}$ and messages
$M_t=\{m_t^a\}_{a\in\mathcal{A}}$ are produced from boundary $t-1$ when
$t>0$ and are applied before boundary $t$ is committed. Let $T_t$ denote the
synchronized engine telemetry at boundary $t$, and let $\mathcal{C}_t$ denote
the combined message, action, and engine receipts. Let $\mathcal{V}_t$ be the
engine-adjudicated event set, $r_t$ the evaluator score, and $d_t$ the
termination flag. The active-tick kernel interface is
\begin{equation}
	\begin{aligned}
		(O_t,I_t,\mathcal{V}_t,r_t,d_t)
		&=W.\operatorname{Step}(U_t,M_t),\\
		(u_{t+1}^a,m_{t+1}^a)
		&=\pi_a.\operatorname{Act}(o_t^a,I_t^a).
	\end{aligned}
	\label{eq:runtime-contract}
\end{equation}
Equation~\ref{eq:runtime-contract} gives the tick order. The kernel first
applies $(U_t,M_t)$ and commits boundary $t$. Each policy then reads its
observation and inbox at that boundary and produces the action and messages for
boundary $t+1$. Reset creates boundary $0$ with empty action and message inputs
before any policy call. Algorithm~1 in main text omits the boundary subscript
for brevity and calls $\mathcal{C}_t$ ``receipts.'' Its per-agent symbols are
the members of the corresponding aggregates. In Algorithm~\ref{alg:step} in appendix here, a
hat marks an action or message admitted after kernel validation, while an
unhatted symbol is the candidate policy output.

\begin{table*}[htpb]
	\centering\small
	\setlength{\tabcolsep}{4pt}
	\caption{Symbols used in Algorithm~1 main text and its operational expansion in Algorithm~\ref{alg:step} in appendix.}
	\label{tab:runtime-symbols}
	\begin{tabular}{@{}p{0.18\textwidth}p{0.76\textwidth}@{}}
		\toprule
		\textbf{Symbol} & \textbf{Definition} \\
		\midrule
		$\mathcal{A},a,\pi_a$ & Set of agents active under $S$, one agent identifier, and its policy. Policy $\pi_a$ reads $(o_t^a,I_t^a)$ and returns $(u_{t+1}^a,m_{t+1}^a)$. \\
		$W=(B,S,G,\mathcal{E},R)$ & World object used by $W.\operatorname{Step}$. Its components are the message bus $B$, reset specification $S$, shared registry $G$, evaluator set $\mathcal{E}$, and episode recorder $R$. Policy calls occur outside $W.\operatorname{Step}$. \\
		$\mathcal{N}_{\mathrm{eng}},q$ & Set of engine nodes participating in $S$ and one node in that set. A boundary commits after every $q\in\mathcal{N}_{\mathrm{eng}}$ supplies a coherent state for tick $t$. \\
		$t$ & Committed boundary index. Boundary $0$ is the synchronized post-reset state. Boundary $t+1$ follows one successful engine advance from $t$. \\
		$K$ & Runtime limits $K=(K_{\mathrm{tick}},K_{\mathrm{model}},K_{\mathrm{engine}})$: maximum committed decision ticks, wall-clock deadline for each policy barrier, and wall-clock deadline for each reset or engine barrier. \\
		$O_t,o_t^a,O_t^{\mathrm{raw}}$ & Policy-visible observation aggregate, agent $a$'s member, and the synchronized unmasked engine observations. Formally $O_t=\{o_t^a\}_{a\in\mathcal{A}}=\operatorname{ProjectVisible}(G,O_t^{\mathrm{raw}},T_t,S)$. \\
		$I_t,I_t^a$ & Aggregate inbox and agent $a$'s inbox, with $I_t=\{I_t^a\}_{a\in\mathcal{A}}$. $I_t^a$ contains committed message edges addressed to $a$ and receipts visible to $a$ under $S$. This defines the aggregate $I$ logged and returned in Algorithm~1. \\
		$u_t^a,m_t^a$ & Agent $a$'s candidate physical command, possibly a no-op, and possibly empty finite list of typed messages. For $t>0$, they are generated from boundary $t-1$ and enter the kernel call that creates boundary $t$. At reset they are empty. \\
		$U_t,M_t$ & Pending candidate aggregates $U_t=\{u_t^a:a\in\mathcal{A}\}$ and $M_t=\{m_t^a:a\in\mathcal{A}\}$. They correspond to $U,M$ in Algorithm~1. \\
		$\widehat U_t,\widehat M_t$ & Admitted results after kernel validation. $\widehat U_t$ contains at most one winning command per controlled resource after schema, authority, and conflict checks. $\widehat M_t$ contains fixed sender-to-recipient edges after envelope validation and topology expansion. \\
		$\mathcal{C}_t,\mathcal{C}_t^M,\mathcal{C}_t^U,\mathcal{C}_t^E$ & Aggregate boundary receipt set and its message-validation, action-validation, and engine subsets. Thus $\mathcal{C}_t=\mathcal{C}_t^M\cup\mathcal{C}_t^U\cup\mathcal{C}_t^E$ is ``receipts'' in Algorithm~1. It contains typed admission, rejection, progress, completion, substitution, and timeout records. \\
		$T_t$ & Typed synchronized engine telemetry, denoted $T$ in Algorithm~1. Its fields and visibility are specified below. \\
		$\mathcal{V}_t$ & Engine-grounded event set returned by \textsc{Adjudicate}, denoted $\mathcal{V}$ in Algorithm~1. \\
		$r_t,d_t$ & Evaluator reward or score and Boolean done flag returned at boundary $t$. The enclosing loop also stops at the tick limit or on an infrastructure abort. \\
		\bottomrule
	\end{tabular}
\end{table*}

In star mode, an agent sends to the coordinator, and a coordinator message uses
its schema-declared recipients. In broadcast mode, a message is routed to all
other active agents. Validation fixes these sender-to-recipient edges before
the bus stages them.

\paragraph{Operator correspondence.}
\textsc{ValidateAndExpand} implements the main algorithm's ``validate and
route'' operation. Its input $M_t$ is the main algorithm's candidate-message
aggregate. Applying the message schema and communication mode in $S$ produces
the admitted edge set $\widehat M_t$ and routing receipts.
\textsc{ValidateAndResolve} implements
``validate and dispatch.'' From candidate aggregate $U_t$, action schema,
authority, registry state, and the deterministic rule below, it produces
$\widehat U_t$ and corresponding validator receipts.

The $B$ \textsc{Stage} operation buffers $\widehat M_t$.
$B$ \textsc{Commit} atomically marks all those edges deliverable only after the
engine barrier. \textsc{DeliveredTo} selects the committed edges for recipient
$a$.
\textsc{AssembleBoundary} returns the pre-projection observation
$O_t^{\mathrm{raw}}$, engine receipts $\mathcal{C}_t^E$, and telemetry $T_t$.
The three receipt subsets form the main algorithm's ``receipts.'' The raw
observation is used by \textsc{Update}, and \textsc{ProjectVisible} produces
the policy-visible $O_t$. Together these operations implement the main
algorithm's compact ``collect $O$, receipts, and $T$'' step.
Clearing each $I_a$ and routing candidate $M_t$ in
Algorithm~1 is implemented by constructing the aggregate
$I_t=\{I_t^a\}$ from committed $\widehat M_t$ edges and receipts in
$\mathcal{C}_t$. \textsc{VisibleReceiptsFor} applies the visibility rules in
$S$. \textsc{Reset} and \textsc{Ready} commit boundary $0$. On later calls,
\textsc{AdvanceTo} implements ``commit one synchronized boundary.''
\textsc{Update}, \textsc{Adjudicate}, \textsc{Evaluate}, and
\textsc{Log} follow the order in Algorithm~1.
\textsc{TerminationReason} reports success, task failure, or budget exhaustion.
An infrastructure abort returns immediately with the partial trace.

\paragraph{Telemetry and visibility.}
$T_t$ is the typed engine-telemetry record corresponding to the main paper's
$T$. It contains the tick, engine and frame IDs, simulation and wall-clock
timestamps, per-asset pose, velocity, resource, and status, sensor frame
IDs and capture timestamps, collisions and contacts, task events, and
engine status.
The separately named receipt set $\mathcal{C}_t$ corresponds to ``receipts'' in
Algorithm~1. Each record contains a request or message ID, source
tick, sender, controlled resource or recipient, status (accepted,
running, completed, rejected, or timed out), reason code, and engine or bus
timestamp.
Completion receipts for an earlier long-running command do not block the
current barrier. A receipt arriving after boundary $t$ is queued and first
appears in $\mathcal{C}_{t+1}$. $O_t^{\mathrm{raw}}$ contains the engine sensor
payloads assembled at that barrier. Hidden target state may occur in $T_t$ for
adjudication and replay, but \textsc{ProjectVisible} removes fields marked
hidden by $S$ before producing $O_t$. Hidden coordinates are omitted from
policy observations and inboxes.

\paragraph{Message timing.}
Validation fixes the recipient set immediately after the policy barrier. A
successful edge in $\widehat M_{t+1}$ is inserted atomically into the
recipient's $I_{t+1}$. Rejected, expired, or disconnected-recipient edges
produce a sender receipt. The message $m_{t+1}^a$, produced while processing
snapshot $t$, is first readable by
broadcast peers in $I_{t+1}$. In star mode the same output is
readable by the coordinator in $I_{t+1}$. A coordinator response produced
after reading it enters $M_{t+2}$ and is first readable by its final recipient in
$I_{t+2}$. Inputs for $I_t$ close when boundary $t$ commits.

\paragraph{Conflict resolution.}
Validation first rejects unknown agents, commands outside the declared action
space, stale tick tags, and senders without control authority. Commands for the
same physical resource retain their order in the collected policy outputs. The
first valid request is admitted, and later requests receive a conflict-rejection
receipt. A completed or rejected request releases the resource for a later tick.

\subsection{Communication Metrics}
\label{app:comm-metrics}

Table~4 in main text is computed directly from the synchronized message and
task-event logs. Star and broadcast use the same rules. Failed and
budget-truncated episodes remain in the matched evaluation set. Infrastructure
failures are rerun with the same instance and seed.

\begin{table*}[htpb]
	\centering\small
	\setlength{\tabcolsep}{4pt}
	\caption{Communication metrics used in Table~4 in main text.}
	\label{tab:comm-definitions}
	\begin{tabular}{@{}p{0.16\textwidth}p{0.76\textwidth}@{}}
		\toprule
		\textbf{Metric} & \textbf{Trace calculation} \\
		\midrule
		Msg & Mean number of logged sender-to-recipient application messages per episode. A broadcast routed to $k$ agents counts as $k$ messages, where $k$ is the number of recipients. Runtime receipts are excluded. \\
		KB & Mean sum of payload bytes for the same logged application messages per episode, divided by $1000$. A broadcast payload is counted once for each recipient. \\
		Rnd & Mean number of synchronized decision ticks containing at least one assignment or target-status message. An episode with no such message contributes zero. \\
		Gap & Mean percentage of the initial requirement that remains incomplete at episode end. The requirement is delivery mass for CD and target count for UFR and SSR. A partial delivery reduces the remaining mass only by its engine-confirmed amount. A pending, rejected, or timed-out action receives no completion credit. The CD demand larger than one vehicle's capacity is handled by the task's allowed split deliveries. A scenario with zero initial requirement is reported as N/A. \\
		Rep & Number of repeated target events divided by all valid target events in the matched episodes. A valid event is an assignment request or target-region entry for a target in the scenario. It is repeated when the target was already assigned or completed. Continuous presence in one target region counts as one entry. If no valid target event occurs, Rep is reported as N/A. \\
		\bottomrule
	\end{tabular}
\end{table*}

\section{Additional Experimental Details}
\subsection{Evaluation Metrics}
\label{app:metrics}

\new{Tables~\ref{tab:metrics-core} and~\ref{tab:metrics-diagnostic} define the metrics used in the main benchmark and diagnostic studies. Unless a subsection states otherwise, each diagnostic condition uses $100$ matched episodes and pools the retained frames, actions, and messages from those traces. The denominator of each rate includes only the episodes or trace elements retained by the stated protocol. Higher values are better unless the definition identifies an error, cost, or delay. For the trajectory analysis in Table~\ref{tab:ctx}, each action sequence is partitioned into three equal intervals by normalized trajectory position before the behavioral rates are computed.}

\begin{table*}[htpb]
	\centering\scriptsize
	\setlength{\tabcolsep}{4pt}
	\caption{\new{Definitions of outcome, coordination, provenance, and perception metrics.}}
	\label{tab:metrics-core}
	\begin{tabular}{@{}p{0.18\textwidth} p{0.74\textwidth}@{}}
		\toprule
		\textbf{Metric} & \textbf{Definition} \\
		\midrule
		\multicolumn{2}{@{}l}{\emph{Task outcome}} \\
		Task success rate (TSR) & Percentage of benchmark episodes in which the simulator confirms every required task terminal condition within the episode budget. The conditions are listed in Table~\ref{tab:task-adjudication}. \\
		Macro average (Avg) & Task-balanced arithmetic mean of TSR across the nine tasks. For a task evaluated under two communication modes, TSR is first averaged across the two modes. \\
		Strict success (SR) & Mean of the binary engine success indicator in a targeted diagnostic study. SR uses the same strict completion principle as TSR, but follows the sample size and infrastructure failure exclusions stated for that diagnostic protocol. \\
		Subtask completion rate (SCR) & Number of required targets, deliveries, rescues, repairs, or other task components completed, divided by the number required in that episode. \\
		\midrule
		\multicolumn{2}{@{}l}{\emph{Coordination}} \\
		Message count & Mean number of logged sender-to-recipient application messages per episode. One broadcast is counted once per recipient. Runtime receipts are excluded. \\
		Payload & Mean payload bytes for the same application messages per episode, divided by $1000$. A broadcast payload is counted once per recipient. \\
		Communication rounds & Mean number of synchronized decision ticks containing an assignment or target-status message. \\
		Assignment gap & Mean percentage of required task amount that remains unconfirmed at episode end. Partial delivery contributes only its remaining amount, and no credit is given without engine confirmation. Lower is better. \\
		Repeated target rate & Number of repeated assignment requests or target-region entries divided by all valid target events. Continuous presence in one region counts once. Lower is better. \\
		\midrule
		\multicolumn{2}{@{}l}{\emph{Subtask provenance}} \\
		Cumulative stage success & Percentage of episodes that complete every task stage up to and including the reported stage. \\
		\midrule
		\multicolumn{2}{@{}l}{\emph{Perception and localization}} \\
		Detection F1 & Harmonic mean of detection precision and recall over aligned frames under the box-matching rule in Sec.~\ref{app:adjudication}, where TP, FP, and FN count true positives, false positives, and false negatives. Precision is $\mathrm{TP}/(\mathrm{TP}+\mathrm{FP})$, and recall is $\mathrm{TP}/(\mathrm{TP}+\mathrm{FN})$. \\
		Miss rate & Number of aligned frames containing at least one unmatched visible ground-truth target, divided by the full set of aligned evaluation frames. Lower is better. \\
		Hallucination rate & Number of aligned frames containing at least one unmatched positive report, divided by the same full set of aligned evaluation frames. Lower is better. \\
		Object confusion rate & Number of evaluated perception calls assigned to an incorrect target class or distractor, divided by all evaluated perception calls. Lower is better. \\
		Localization error (LocErr) & Median $d_{xy}(\widehat p,p^*)$ over box-matched true-positive reports that contain a world-coordinate prediction. \\
		Expected calibration error (ECE) & Mean absolute difference between box-match accuracy and mean confidence across ten bins of equal width, weighted by the number of samples in each bin. Lower is better. \\
		Attention IoU & Intersection over union between the ground truth target box and the image attention region obtained by thresholding attention at its mean plus one standard deviation. \\
		F1 domain difference & Let $\mathrm{F1}_{\mathrm{sim}}$ and $\mathrm{F1}_{\mathrm{real}}$ denote F1 on matched simulator and GPT-generated real-style images. Their difference is $\mathrm{F1}_{\mathrm{real}}-\mathrm{F1}_{\mathrm{sim}}$. A negative value indicates lower performance in the real-style images. \\
		\bottomrule
	\end{tabular}
\end{table*}

\begin{table*}[htpb]
	\centering\scriptsize
	\setlength{\tabcolsep}{4pt}
	\caption{\new{Definitions of failure, behavior, efficiency, and trace metrics.}}
	\label{tab:metrics-diagnostic}
	\begin{tabular}{@{}p{0.18\textwidth} p{0.74\textwidth}@{}}
		\toprule
		\textbf{Metric} & \textbf{Definition} \\
		\midrule
		\multicolumn{2}{@{}l}{\emph{Failure and behavior}} \\
		Failure family share & Percentage of retained failed episodes whose earliest breakdown supported by the trace belongs to a given failure family. Infrastructure failures are excluded and each episode receives one label. \\
		Constraint adherence & Number of evaluated actions that satisfy all active task and motion constraints, divided by all evaluated actions. \\
		Repetition rate & Number of actions whose normalized action signature matches the immediately preceding action from the same agent, divided by all actions after the first action in each sequence. Lower is better. \\
		Out of bounds rate & Number of evaluated actions that leave the configured operating region or violate an active altitude or motion bound, divided by all evaluated actions. Lower is better. \\
		Forgotten subgoal rate & Number of evaluated actions that omit an active requirement or act on an already completed subgoal, divided by all evaluated actions. Lower is better. \\
		\midrule
		\multicolumn{2}{@{}l}{\emph{Efficiency and output validity}} \\
		Inference token cost & Total runtime prompt and completion tokens used by model calls. Relative cost divides this total by the designated full agent or prompt baseline. \\
		JSON compliance & Number of model outputs that satisfy the required JSON schema, divided by the number of evaluated model calls. \\
		\midrule
		\multicolumn{2}{@{}l}{\emph{Search and execution traces}} \\
		Search coverage & Number of distinct valid search cells visited, divided by the number of cells in the configured search region. \\
		First discovery step & Decision step at which the first valid target detection is recorded. Lower is better. \\
		Redundant revisit rate & Number of search visits to cells that were already covered, divided by all search visits. Lower is better. \\
		Idle fraction & Number of agent steps with no assigned or executed task action, divided by all agent steps. Lower is better. \\
		Communication at discovery events & Number of messages aligned with a validated target discovery event, divided by all evaluated messages. \\
		\bottomrule
	\end{tabular}
\end{table*}

\subsection{Sea-Rescue Communication Analysis}
\label{app:sea-comm}

\begin{figure*}[htpb]
	\centering
	\begin{minipage}[t]{0.49\textwidth}
		\centering
		\includegraphics[width=\linewidth]{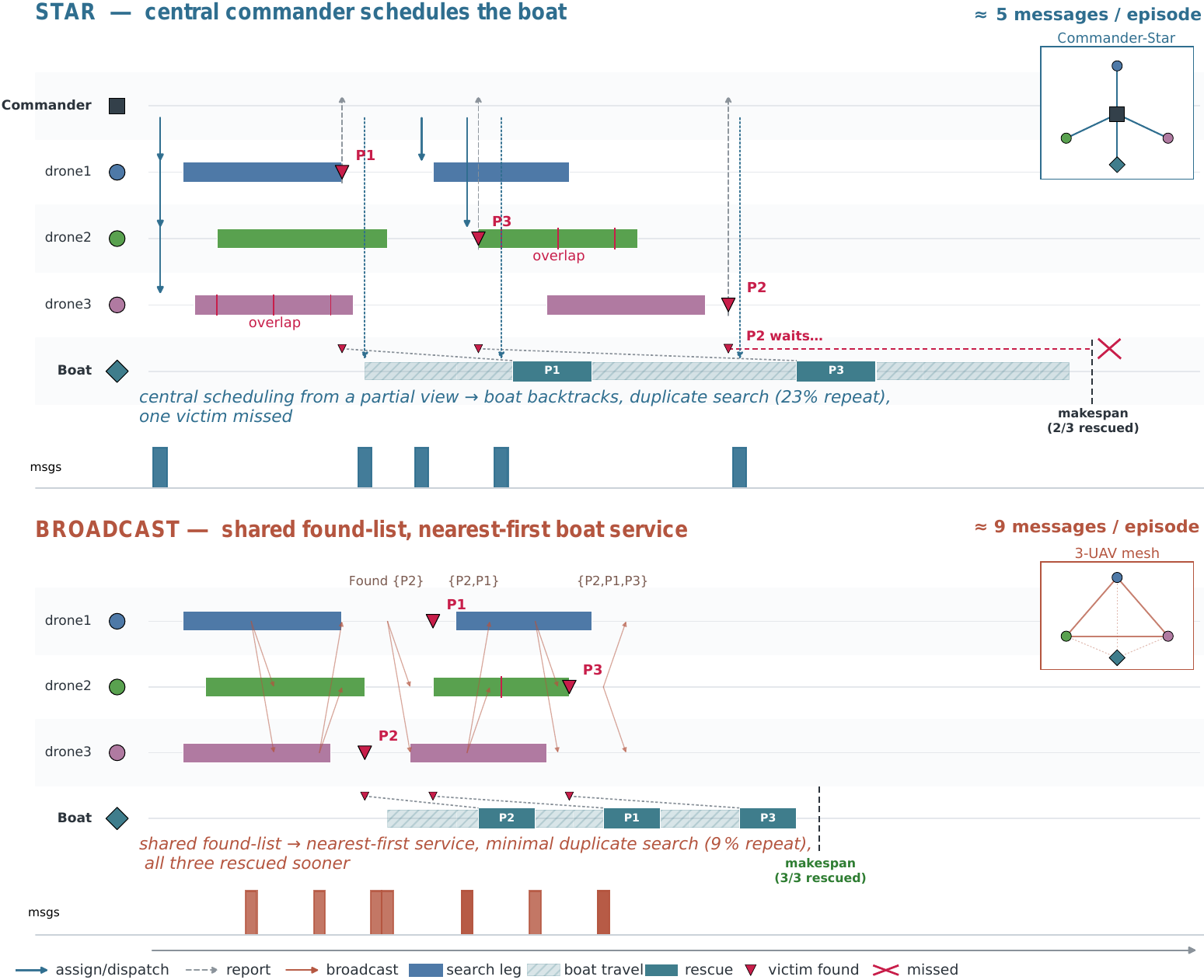}
		\\[-0.4em]\footnotesize (a)
	\end{minipage}
	\hfill
	\begin{minipage}[t]{0.49\textwidth}
		\centering
		\includegraphics[width=\linewidth]{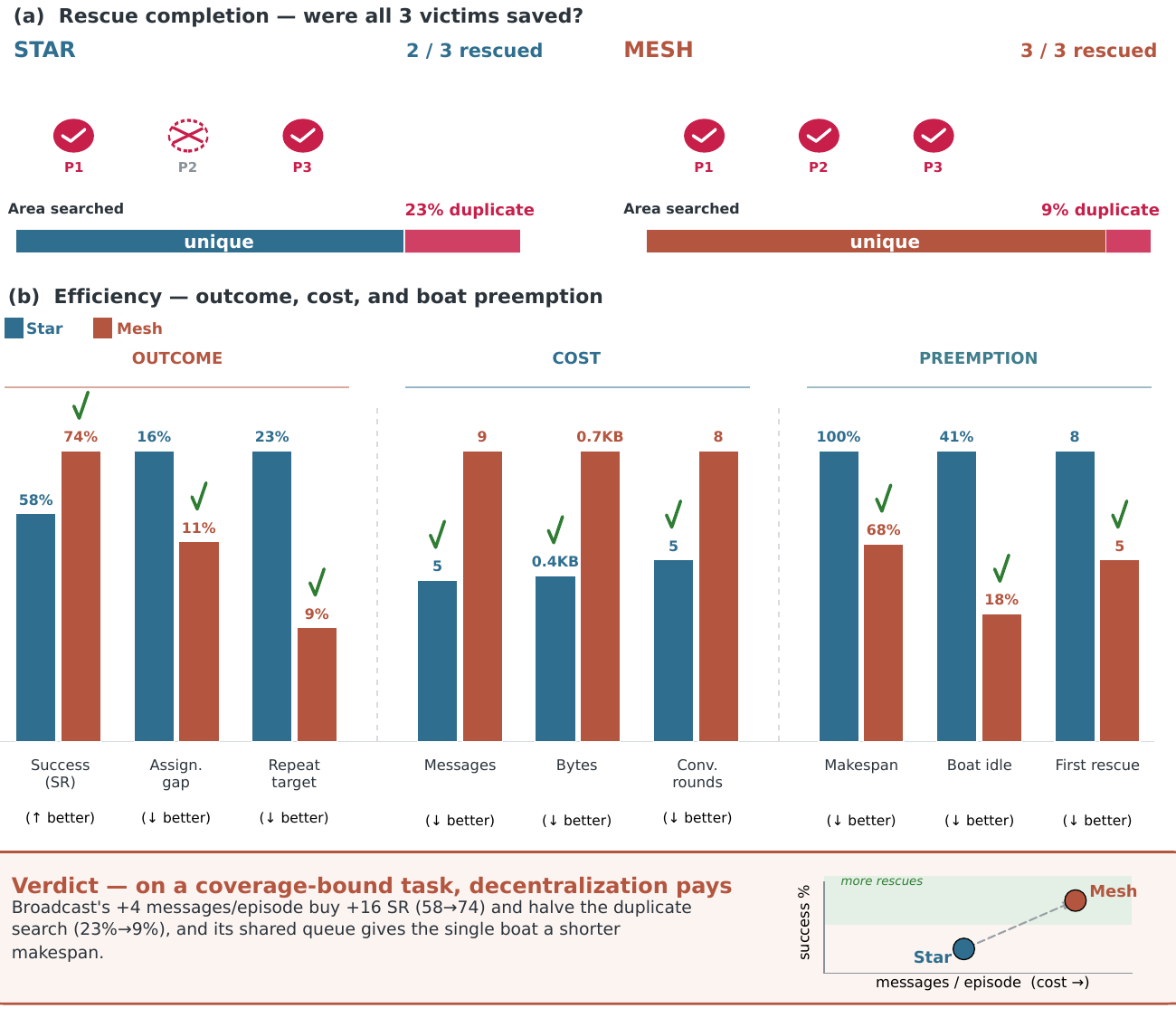}
		\\[-0.4em]\footnotesize (b)
	\end{minipage}
	\caption{\new{Sea-rescue comparison with fixed qwen3.7-plus over $100$ matched instances. (a) Representative episode, where star rescues two of three victims and broadcast rescues all three. (b) Matched-instance outcomes, where broadcast increases TSR and lowers the unconfirmed-demand and duplicate-target indicators.}}
	\label{fig:sea-process}
\end{figure*}

Figure~\ref{fig:sea-process} supplements Table~4 in main text with a representative sea-rescue episode and outcomes over $100$ matched instances under fixed qwen3.7-plus. Broadcast rescues all three victims in the shown episode, compared with two under star. Across matched instances, it raises TSR from $58\%$ to $74\%$ and lowers the unconfirmed-demand and duplicate-target indicators from $16\%$ to $11\%$ and from $23\%$ to $9\%$, while increasing messages from $5$ to $9$, payload from $0.4$ to $0.7$ KB, and communication rounds from $5$ to $8$.

\subsection{Trace-Based Behavioral Analysis}
\label{sec:exp:viz}
\new{Table~\ref{tab:viz} summarizes trajectory, perception, and communication indicators for GPT-5.5 and Gemma-3-4B over $100$ matched sea-rescue (broadcast) episodes.}
\begin{table}[htpb]
	\centering\footnotesize
	\setlength{\tabcolsep}{2pt}
	\caption{\new{Behavioral signatures over $100$ matched sea-rescue (broadcast) episodes. All values are arithmetic means of per-episode statistics. First discovery is averaged over episodes with a valid target detection and is reported in decision steps. Coverage, redundant revisits, and event-aligned communication are percentages, and idle fraction is in $[0,1]$.}}
	\label{tab:viz}
	\begin{tabular}{@{}L{0.34\linewidth} rr L{0.24\linewidth}@{}}
		\toprule
		\textbf{Signature} & \textbf{GPT-5.5} & \textbf{Gemma} & \textbf{What it reveals} \\
		\midrule
		Search coverage        & \textbf{88} & 48 & completeness of the sweep \\
		First discovery step   & \textbf{12} & 34 & steps to first target \\
		Redundant revisits     & \textbf{9}  & 38 & wasted re-scans from drift \\
		Idle-agent fraction    & \textbf{0.12} & 0.45 & load balance and coordination \\
		Comm.\ at discovery   & \textbf{82} & 33 & messages tied to events \\
		\bottomrule
	\end{tabular}
\end{table}

\new{\textbf{The traces expose different execution patterns (Table~\ref{tab:viz}).} GPT-5.5 searches more broadly, reaches targets earlier, and revisits fewer cells than Gemma-3-4B. It also has a lower idle fraction and more messages aligned with discovery events.}

\subsection{Failure Analysis}
\label{sec:exp:fa}

\new{Failure attribution uses $100$ matched episodes per condition. Each failed episode is assigned to its earliest model-caused breakdown using aligned perception, localization, coordination, action, and task-state traces. Excluding infrastructure failures leaves $2{,}515$ API-model and $5{,}381$ open-weight-model failures. Shares are computed within each model and task condition. Star and broadcast are averaged within UFR, SSR, and CD before the unweighted average across tasks and models in each tier.}

\begin{table}[htpb]
	\centering\scriptsize
	\setlength{\tabcolsep}{4pt}
	\caption{\new{Task- and model-balanced macro-average failure attribution over retained model-caused failed episodes, where $n$ is the number of failed episodes, with API $n=2{,}515$ and open $n=5{,}381$. Each episode is assigned to the earliest supported failure family. Rows sum to $100\%$ within a tier.}}
	\label{tab:fa}
	\begin{tabular}{@{}l rr l@{}}
		\toprule
		\textbf{Failure family} & \textbf{API} & \textbf{Open} & \textbf{Typical task} \\
		\midrule
		Perception miss (missed detection)   & 30 & 42 & sea, pipeline \\
		Object confusion (misrecognition)    & 12 & 23 & sea, fire \\
		Localization error (outside task radius) & 16 & 14 & fire, pipeline \\
		Long-context drift                   & 18 & 13 & logistics, factory fire \\
		Coordination deadlock                & 14 & 5  & logistics, fire \\
		Handoff or dependency break           & 10 & 3  & pipeline, fire \\
		\bottomrule
	\end{tabular}
\end{table}

\new{\textbf{Where failures occur (Table~\ref{tab:fa}).} Missed detection and object confusion account for $42\%$ of the task- and model-balanced API distribution and $65\%$ of the open-weight distribution. Coordination deadlock and handoff failure account for $24\%$ and $8\%$, respectively.}

\begin{table}[htpb]
	\centering\small
	\setlength{\tabcolsep}{4.5pt}
	\caption{\new{Perception indicators by backbone (\%, except IoU): miss, false-positive (Halluc.), object-confusion, and attention-to-target IoU. Attention maps are available only for the evaluated open checkpoints. Unavailable API-model values are marked N/A.}}
	\label{tab:perc}
	\begin{tabular}{@{}l rrr r@{}}
		\toprule
		\textbf{Backbone} & \textbf{Miss} & \textbf{Halluc.} & \textbf{Confus.} & \textbf{Attn-IoU} \\
		\midrule
		\multicolumn{5}{@{}l}{\emph{Commercial models}} \\
		GPT-5.5        & \textbf{28} & 2 & \textbf{8}  & N/A \\
		GPT-5.4        & 38 & 2 & 12 & N/A \\
		Gemini-3.1-Pro & 32 & \textbf{1} & 10 & N/A \\
		Kimi-K2.5      & 46 & 3 & 18 & N/A \\
		Qwen3-VL-Flash & 52 & 3 & 22 & N/A \\
		\midrule
		\multicolumn{5}{@{}l}{\emph{Open-source models}} \\
		Qwen3-VL-2B    & 40 & 4 & 24 & 0.29 \\
		Qwen3-VL-4B    & 55 & 4 & 26 & 0.24 \\
		Qwen3-VL-8B    & 44 & 3 & 20 & \textbf{0.34} \\
		InternVL3.5-8B & 47 & 3 & 21 & 0.31 \\
		MiniCPM-V-4.5  & 62 & 5 & 30 & 0.19 \\
		Gemma-3-4B     & 66 & 6 & 33 & 0.17 \\
		Pixtral-12B    & 50 & 4 & 25 & 0.27 \\
		\bottomrule
	\end{tabular}
\end{table}

\new{\textbf{What accompanies perception failure (Table~\ref{tab:perc}).} Errors are dominated by omissions. The false-positive rate ranges from $1\%$ to $6\%$, while the miss rate ranges from $28\%$ for GPT-5.5 to $66\%$ for Gemma-3-4B and is highest for small water victims and pipe leaks. Among open checkpoints, higher attention-to-target IoU is associated with lower confusion. Qwen3-VL-8B has the highest IoU ($0.34$) and a $20\%$ confusion rate. Gemma-3-4B has the lowest IoU ($0.17$) and a $33\%$ confusion rate. Trace examples place attention on wave glint, buoys, glare, and wet ground.}

\begin{table}[htpb]
	\centering\small
	\setlength{\tabcolsep}{5pt}
	\caption{\new{Behavioral indicators by trajectory tercile (\%). Ctx. denotes trajectory interval, Adher. constraint adherence, Repeat action repetition, OOB out-of-bounds actions, and Forgot forgotten subgoals. Adherence is higher-is-better, while the other indicators are error rates.}}
	\label{tab:ctx}
	\begin{tabular}{@{}ll rrrr@{}}
		\toprule
		& \textbf{Ctx.} & \textbf{Adher.} & \textbf{Repeat} & \textbf{OOB} & \textbf{Forgot} \\
		\midrule
		Commercial & Early & 94 & 8  & 3  & 4  \\
		& Mid   & 88 & 14 & 6  & 9  \\
		& Late  & 79 & 22 & 11 & 17 \\
		\midrule
		Open       & Early & 88 & 14 & 7  & 9  \\
		& Mid   & 74 & 27 & 15 & 21 \\
		& Late  & 58 & 41 & 26 & 34 \\
		\bottomrule
	\end{tabular}
\end{table}

\new{\textbf{Later trajectory stages are associated with drift (Table~\ref{tab:ctx}).} From the first to the final trajectory tercile, constraint adherence falls and repeated search, out-of-bounds actions, and forgotten subgoals rise. The adherence drop is $30$ points for open-weight models ($88\%{\to}58\%$) and $15$ points for API models ($94\%{\to}79\%$). Typical late-stage errors revisit searched cells, violate altitude limits, or target already-rescued entities.}

\subsection{Ablation Study}
\label{sec:exp:ablation}

\new{We ablate one agent component at a time on sea rescue (broadcast) and fire response (broadcast), holding the \texttt{qwen3.7-plus} model, scenario, and seed schedule fixed and excluding infrastructure failures. Table~\ref{tab:abl} reports strict success (SR), subtask completion rate (SCR), median localization error, and inference-token cost relative to the full agent. Table~\ref{tab:abl-llm} separately varies the decision layer across an API and an open-weight backbone. Sensing parameters are swept in Sec.~5.5.}

\begin{table}[htpb]
	\centering\small
	\setlength{\tabcolsep}{4.5pt}
	\caption{\new{Component ablation (\texttt{qwen3.7-plus}), excluding infrastructure failures. Each row removes or coarsens one component. SR and SCR are percentages. LocErr is in meters. Cost denotes relative inference tokens.}}
	\label{tab:abl}
	\begin{tabular}{@{}l rrr r@{}}
		\toprule
		\textbf{Variant} & \textbf{SR} & \textbf{SCR} & \textbf{LocErr} & \textbf{Cost}$\times$ \\
		\midrule
		\multicolumn{5}{@{}l}{\emph{Sea rescue (broadcast)}} \\
		Full agent                       & \textbf{74} & \textbf{82} & \textbf{52}  & 1.0 \\
		Single-pass search               & 62 & 70 & 52  & 0.7 \\
		Top view only                    & 58 & 66 & 88  & 0.95 \\
		Heuristic sweep                  & 20 & 34 & 130 & 0.0 \\
		Coarse grid $2^2$ (vs $3^2$)     & 64 & 73 & 52  & 0.25 \\
		\midrule
		\multicolumn{5}{@{}l}{\emph{Fire response (broadcast)}} \\
		Full agent (re-search on)        & \textbf{60} & \textbf{88} & 55 & 1.0 \\
		Single-pass search               & 48 & 67 & 55 & 0.30 \\
		\bottomrule
	\end{tabular}
\end{table}

\new{\textbf{Which components matter (Table~\ref{tab:abl}).} In sea rescue, replacing the LLM decision layer with the heuristic sweep produces the largest observed reduction in SR ($74\%{\to}20\%$) and raises localization error from $52$ to $130$\,m. Removing the front view lowers SR by $16$ points and raises localization error to $88$\,m, indicating that the second view provides useful confirmation for small targets. Bounded re-search raises sea-rescue SR from $62\%$ to $74\%$ and fire-response SCR from $67\%$ to $88\%$. Relative to a single pass, it increases inference tokens by about $1.4\times$ in sea rescue and $3.3\times$ in fire response. Coarsening the sea-rescue grid reduces relative token cost to $0.25\times$ and SR to $64\%$.}

\begin{table}[htpb]
	\centering\small
	\setlength{\tabcolsep}{6pt}
	\caption{\new{Net effect of the LLM decision layer across backbones on sea rescue (broadcast): full agent versus the heuristic (no-LLM) fallback. SR and SCR are percentages. $\Delta\mathrm{SR}$ is the SR difference from the heuristic.}}
	\label{tab:abl-llm}
	\begin{tabular}{@{}l rr r@{}}
		\toprule
		\textbf{Decision backbone} & \textbf{SR} & \textbf{SCR} & \textbf{$\Delta\mathrm{SR}$} \\
		\midrule
		Heuristic (no LLM)   & 20 & 34 & N/A \\
		Qwen3-VL-8B (open)   & 59 & 69 & $+39$ \\
		GPT-5.5 (commercial) & \textbf{84} & \textbf{88} & $+64$ \\
		\bottomrule
	\end{tabular}
\end{table}

\new{\textbf{Decision-layer gains depend on the backbone (Table~\ref{tab:abl-llm}).} The model-free heuristic provides an SR of $20\%$. Under the same sea-rescue scaffold, Qwen3-VL-8B adds $39$ points and GPT-5.5 adds $64$ points.}

\subsection{Photographic Appearance Gap}
\label{sec:exp:sim2real}

\begin{figure*}[htpb]
	\centering
	\begin{minipage}[t]{0.32\textwidth}
		\centering
		\textbf{Fire}\par\smallskip
		\includegraphics[width=\linewidth]{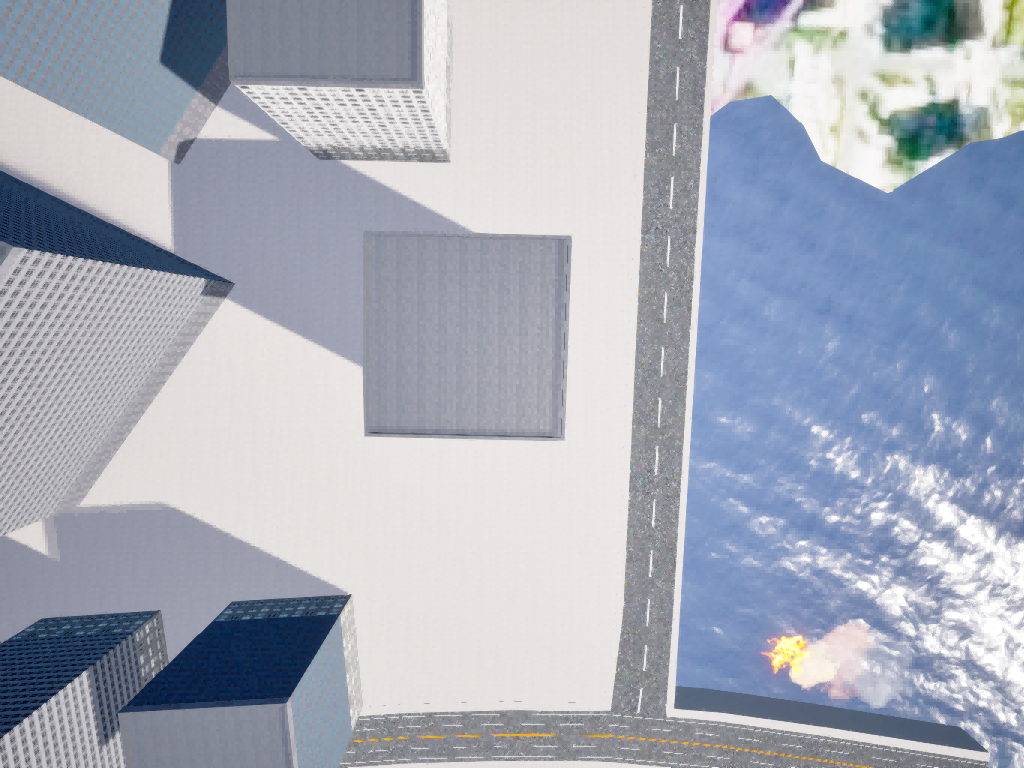}
		\\\footnotesize Simulator frame\par\smallskip
		\includegraphics[width=\linewidth]{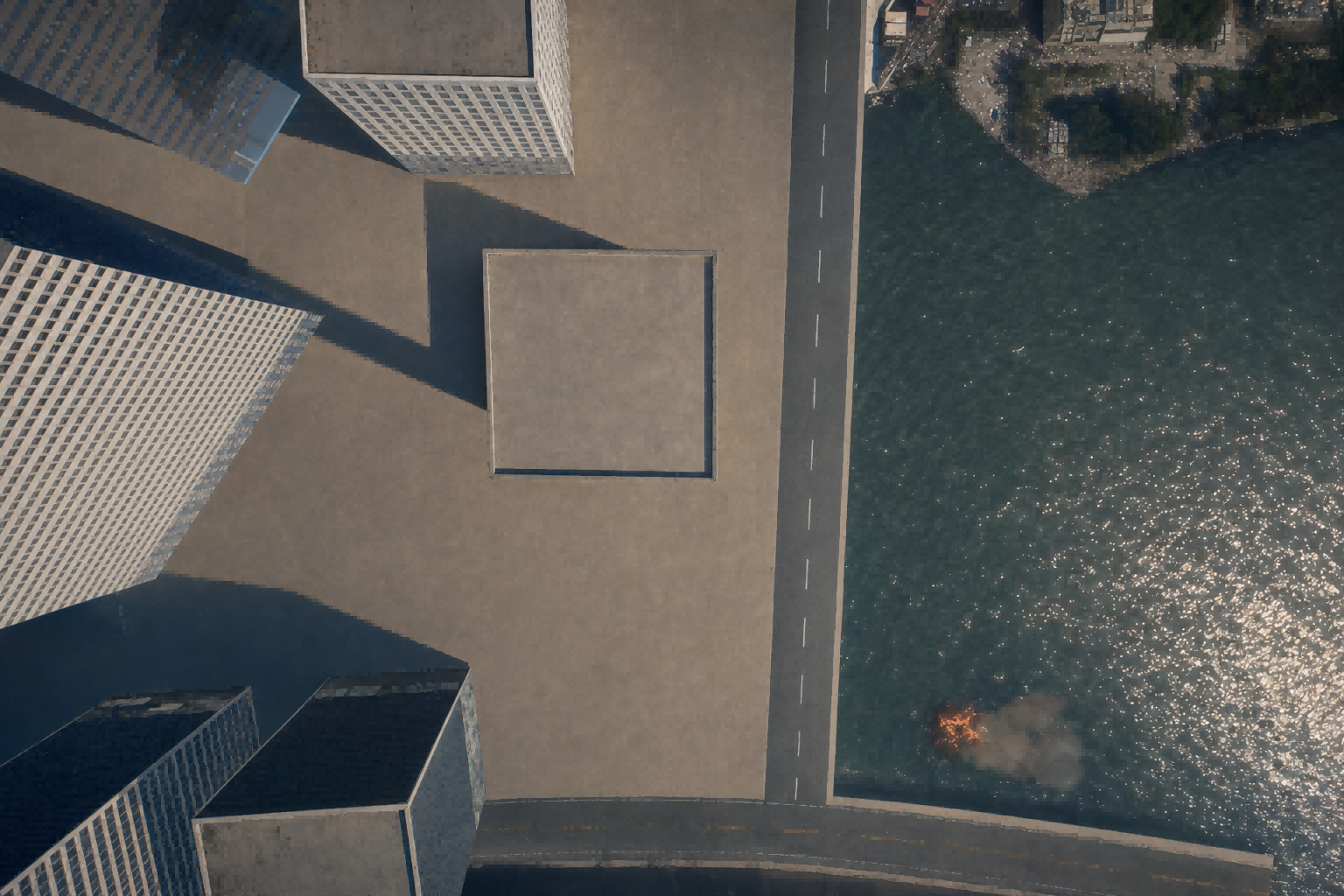}
		\\\footnotesize GPT-generated real-style image
	\end{minipage}
	\hfill
	\begin{minipage}[t]{0.32\textwidth}
		\centering
		\textbf{Pipe leak}\par\smallskip
		\includegraphics[width=\linewidth]{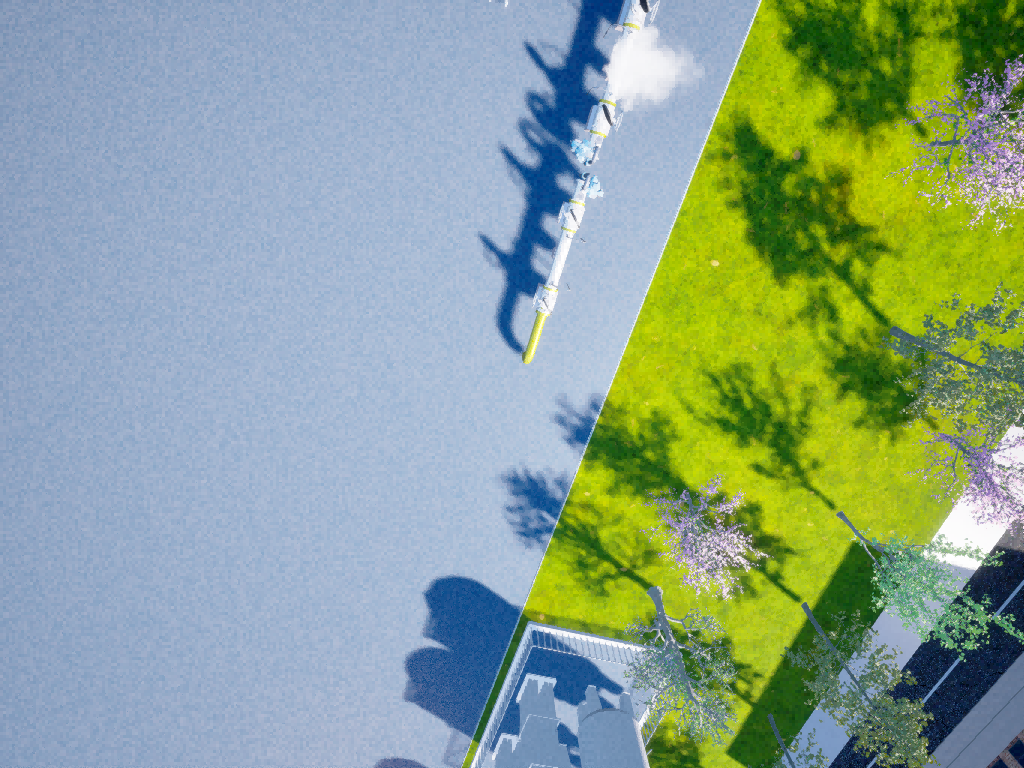}
		\\\footnotesize Simulator frame\par\smallskip
		\includegraphics[width=\linewidth]{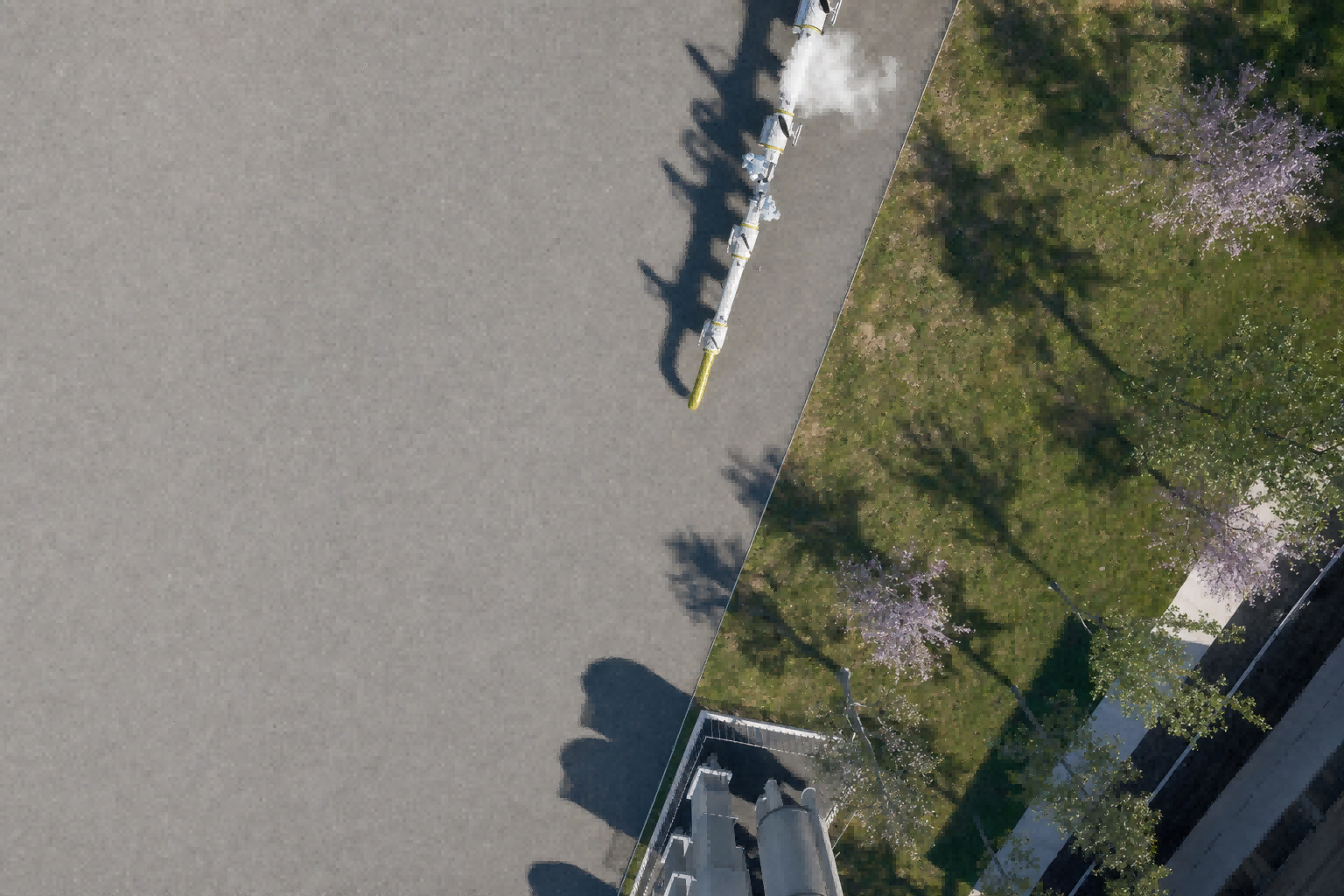}
		\\\footnotesize GPT-generated real-style image
	\end{minipage}
	\hfill
	\begin{minipage}[t]{0.32\textwidth}
		\centering
		\textbf{Water victim}\par\smallskip
		\includegraphics[width=\linewidth]{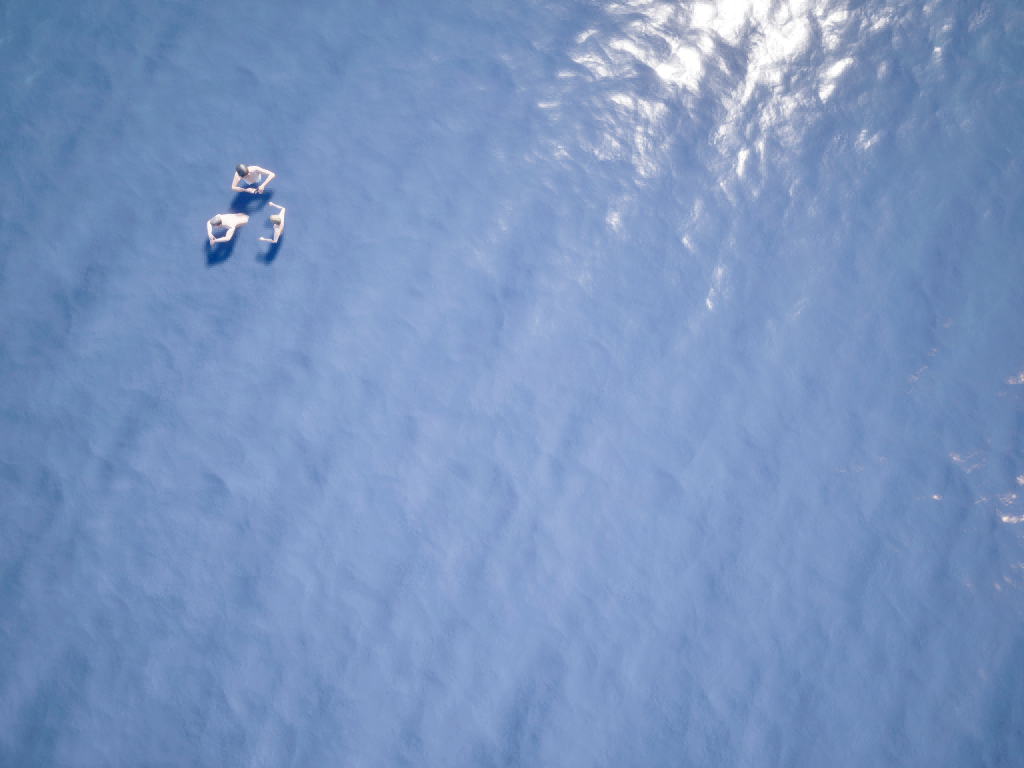}
		\\\footnotesize Simulator frame\par\smallskip
		\includegraphics[width=\linewidth]{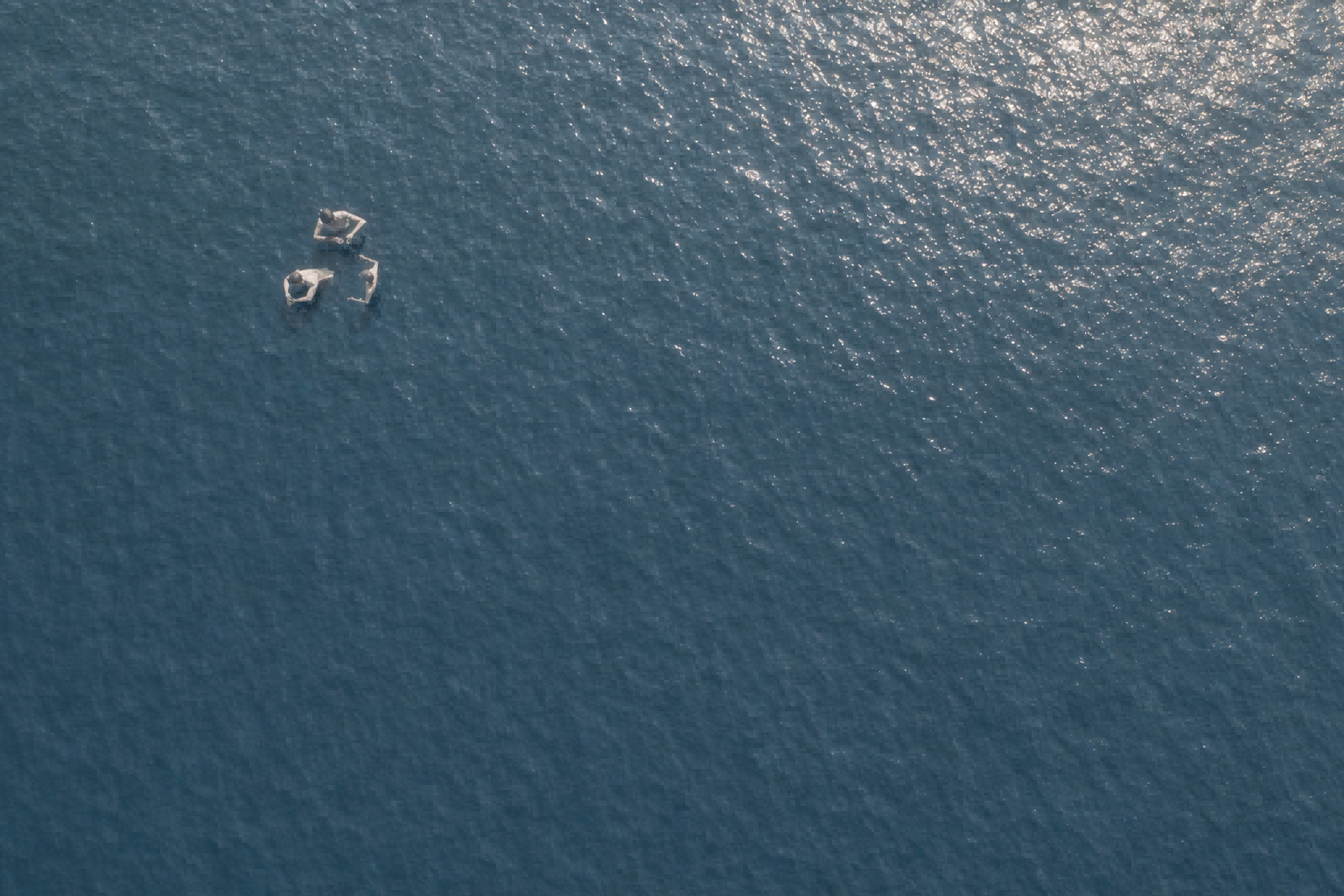}
		\\\footnotesize GPT-generated real-style image
	\end{minipage}
	\caption{\new{Paired examples from the photographic-appearance probe. Each column shows a Lingjing simulator frame above its GPT-generated real-style counterpart.}}
	\label{fig:appearance-pairs}
\end{figure*}

\new{We measure sensitivity to photographic appearance with an offline detection probe. For each category, source frames are sampled from $100$ matched seeded Lingjing episodes and edited with \texttt{gpt-image-2}. The resulting synthetic real-style domain uses the subscript ``real.'' The edit preserves camera view, scene geometry, target presence, count, and relative location while changing visual appearance. Both members of each pair use the same image preprocessing. Figure~\ref{fig:appearance-pairs} shows three examples.}

\new{Within each category, the simulator frames contain a balanced mixture of target-present cases and hard negatives. Each real-style image inherits its source label. Edits that alter target presence or count are rejected. GPT-5.5 and Qwen3-VL-8B process both domains using the same category-specific detection prompt and fixed binary response schema. A target-present image predicted yes is a true positive (TP). A target-absent image predicted yes is a false positive (FP). A target-present image predicted no is a false negative (FN). Counts are pooled by model, category, and domain. We compute $\mathrm{F1}=2\mathrm{TP}/(2\mathrm{TP}+\mathrm{FP}+\mathrm{FN})$ on the $[0,1]$ scale and $\Delta=\mathrm{F1}_{\mathrm{real}}-\mathrm{F1}_{\mathrm{sim}}$.}

\begin{table}[htpb]
	\centering\small
	\setlength{\tabcolsep}{4pt}
	\caption{\new{F1 on Lingjing simulator frames and GPT-generated real-style images. $\Delta=\mathrm{F1}_{\mathrm{real}}-\mathrm{F1}_{\mathrm{sim}}$.}}
	\label{tab:sim2real}
	\resizebox{0.6\columnwidth}{!}{%
		\begin{tabular}{@{}l rr r rr r@{}}
			\toprule
			& \multicolumn{3}{c}{Commercial (GPT-5.5)} & \multicolumn{3}{c}{Open (Qwen3-VL-8B)} \\
			\cmidrule(lr){2-4}\cmidrule(lr){5-7}
			\textbf{Category} & sim & real & $\Delta$ & sim & real & $\Delta$ \\
			\midrule
			Fire (large)          & 0.90 & 0.82 & $-$0.08 & 0.78 & 0.60 & $-$0.18 \\
			Vehicle or crowd      & 0.88 & 0.78 & $-$0.10 & 0.74 & 0.55 & $-$0.19 \\
			Bridge defect         & 0.74 & 0.58 & $-$0.16 & 0.60 & 0.38 & $-$0.22 \\
			Pipe leak (small)     & 0.62 & 0.40 & $-$0.22 & 0.46 & 0.24 & $-$0.22 \\
			Water victim (small)  & 0.58 & 0.33 & $-$0.25 & 0.43 & 0.20 & $-$0.23 \\
			\bottomrule
		\end{tabular}
	}
\end{table}

\new{Table~\ref{tab:sim2real} shows lower F1 in the generated real-style domain for both models in every category. GPT-5.5 drops by $0.08$ to $0.10$ on fire and vehicle or crowd targets, compared with $0.22$ to $0.25$ on pipe leaks and water victims. Qwen3-VL-8B follows the same pattern and reaches $0.20$ to $0.24$ F1 on the two small-target categories. Small, low-contrast targets show the largest decline under this appearance change.}

\subsection{Last-Mile Failure Diagnosis}
\label{app:lastmile}

\begin{figure*}[htpb]
	\centering
	\includegraphics[width=\textwidth]{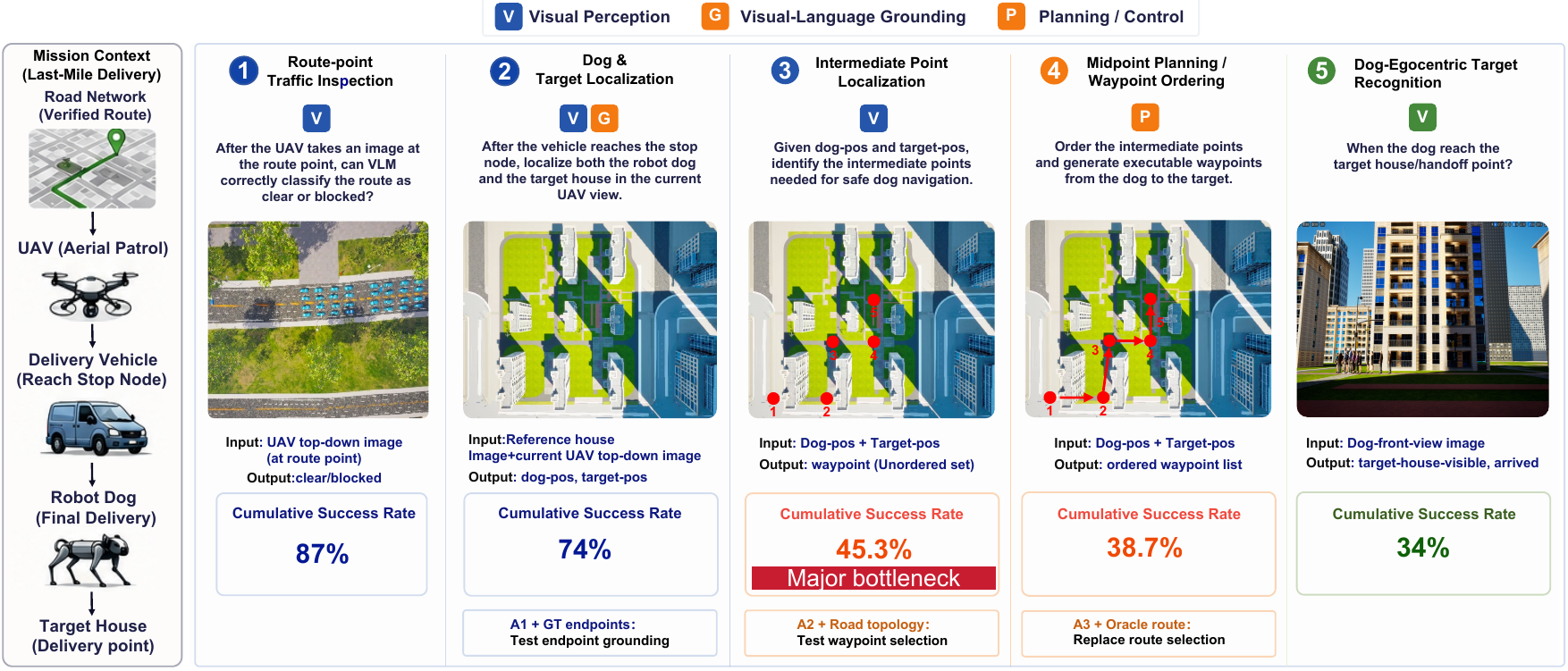}
	\caption{Last-mile provenance with fixed \texttt{qwen3.7-plus} over $150$ episodes. S1 through S5 denote aerial route inspection, endpoint grounding, waypoint generation, legal-edge arrival, and target-house recognition. Percentages report cumulative success through each stage.}
	\label{fig:prov-case}
\end{figure*}

The cumulative chain has five stages. S1 records successful aerial route inspection, and S2 records localization of the route start and target endpoints. S3 requires intermediate waypoint generation on the road graph. S4 requires legal-edge arrival at the vehicle stop node, and S5 requires recognition of the target house after robot-dog deployment. Figure~\ref{fig:prov-case} traces this chain over $150$ episodes. Cumulative success from S1 through S5 is $87\%$, $74\%$, $45.3\%$, $38.7\%$, and $34\%$. The largest loss is the $28.7$-point drop from S2 to S3 when localized endpoints must be converted into intermediate waypoints. Waypoint ordering and road-compliant execution then reduce success by $6.6$ points, followed by a $4.7$-point loss in target recognition.

\begin{table}[htpb]
	\centering\footnotesize
	\setlength{\tabcolsep}{3pt}
	\caption{Last-mile cumulative S4 route-arrival success over $150$ matched episodes under staged interventions. A0 to A2 use \texttt{qwen3.7-plus}, while A3 uses $A^{\star}$ shortest-path search for deterministic next-node selection on the road graph.}
	\label{tab:lastmile}
	\resizebox{0.6\columnwidth}{!}{%
		\begin{tabular}{@{}l l r@{}}
			\toprule
			\textbf{Condition} & \textbf{Information supplied} & \textbf{S4} \\
			\midrule
			A0 Full policy & Images, no GT geometry & 38.7\% \\
			A1 +GT endpoints & GT start and target positions & 72.7\% \\
			A2 +Graph and route guidance & Endpoints, graph, and route hint & 87.3\% \\
			A3 +Shortest path & $A^{\star}$ next-node selection & 94.7\% \\
			\bottomrule
		\end{tabular}
	}
\end{table}

Table~\ref{tab:lastmile} measures every intervention at the cumulative S4
endpoint. Success rises from $58/150$ under A0 to $109/150$ with ground-truth
endpoints, $131/150$ with graph and route guidance, and $142/150$ with
deterministic $A^{\star}$ selection. The respective gains are $51$, $22$, and
$11$ episodes. These S4 diagnostic rates are distinct from the full-task TSRs
in main text Table~3.

\subsection{Prompt Specification for Urban Embodied Agents}
\label{sec:exp:prompt}

\begin{figure}[htpb]
	\centering
	\includegraphics[width=0.5\linewidth]{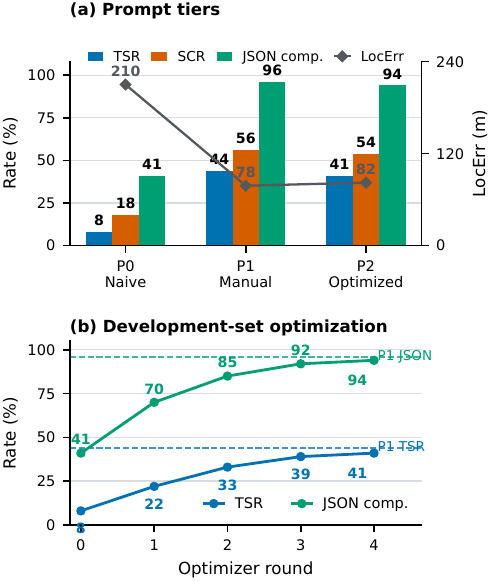}
	\caption{\new{Prompt ablation on broadcast sea rescue with Qwen3-VL-8B. (a) Held-out TSR, SCR, JSON compliance, and LocErr over $100$ matched episodes. (b) Development-set TSR and JSON compliance across four P2 revisions. Higher rates and lower LocErr are better.}}
	\label{fig:prompt}
\end{figure}

\new{We evaluate prompt specification on broadcast sea rescue with Qwen3-VL-8B, fixing the scenario, decoding, action schema, and $100$ matched evaluation episodes. \emph{P0} is a one-line instruction. \emph{P1} adds a perception checklist, coordinate frame, decision and motion rules, and a JSON output contract. Starting from P0, \emph{P2} undergoes four trace-guided revisions. GPT-5.5 proposes each revision from failed Qwen3-VL-8B traces and retains the best candidate on a fixed, disjoint development set. Figure~\ref{fig:prompt}(b) reports this development trajectory. The final comparison of P0 and P2 uses the held-out matched evaluation episodes.}

\new{TSR, SCR, and LocErr follow the preceding definitions. Here TSR requires engine-confirmed rescue of every victim, SCR averages the fraction rescued per episode, and LocErr is the median Euclidean error over matched true-positive locations. JSON compliance is the fraction of outputs that satisfy the required schema.}

\new{Figure~\ref{fig:prompt} shows that prompt specification affects both parseability and task outcomes. Relative to P0, P1 raises TSR from $8\%$ to $44\%$, SCR from $18\%$ to $56\%$, and JSON compliance from $41\%$ to $96\%$, while reducing LocErr from $210$ to $78$\,m. After four revisions, P2 reaches $41\%$ TSR, $54\%$ SCR, $94\%$ JSON compliance, and $82$\,m LocErr. Traces show fewer format failures and reversed-direction actions after introducing explicit schemas and coordinate conventions. P1's $96\%$ JSON compliance and $44\%$ TSR show that format validity alone does not determine task completion.}

\end{document}